\documentclass[conference]{IEEEtran}
\usepackage{times}

\usepackage[numbers]{natbib}
\usepackage{multicol}
\usepackage[bookmarks=true]{hyperref}

\let\labelindent\relax

\usepackage{booktabs}
\usepackage{float}
\usepackage{graphicx}
\usepackage{amsfonts}
\usepackage{bbding}
\usepackage{multirow}
\usepackage{makecell}
\usepackage{caption}
\usepackage{amsmath}
\usepackage{xcolor,colortbl}
\usepackage{enumitem}
\usepackage{mathrsfs}
\usepackage{hyperref}

\definecolor{tablecolor}{RGB}{240, 220, 220}

\begin{document}

\title{SKIL: Semantic Keypoint Imitation Learning for Generalizable Data-efficient Manipulation}


\author{Shengjie Wang$^{1,2,3}$ \  Jiacheng You$^{1,2,3}$ \  Yihang Hu$^{1}$ \  Jiongye Li$^{4}$ \ Yang Gao$^{1,2,3,\dagger}$  \\
$^{1}$IIIS, Tsinghua University \quad $^{2}$Shanghai AI Laboratory \quad  \\ $^{3}$Shanghai Qi Zhi Institute \quad $^{4}$Department of Automation, Tsinghua University \\
}



%

\twocolumn[{%
\renewcommand\twocolumn[1][]{#1}%
\maketitle
\begin{center}
    \centering
    \captionsetup{type=figure}
    \includegraphics[width=0.9\textwidth]{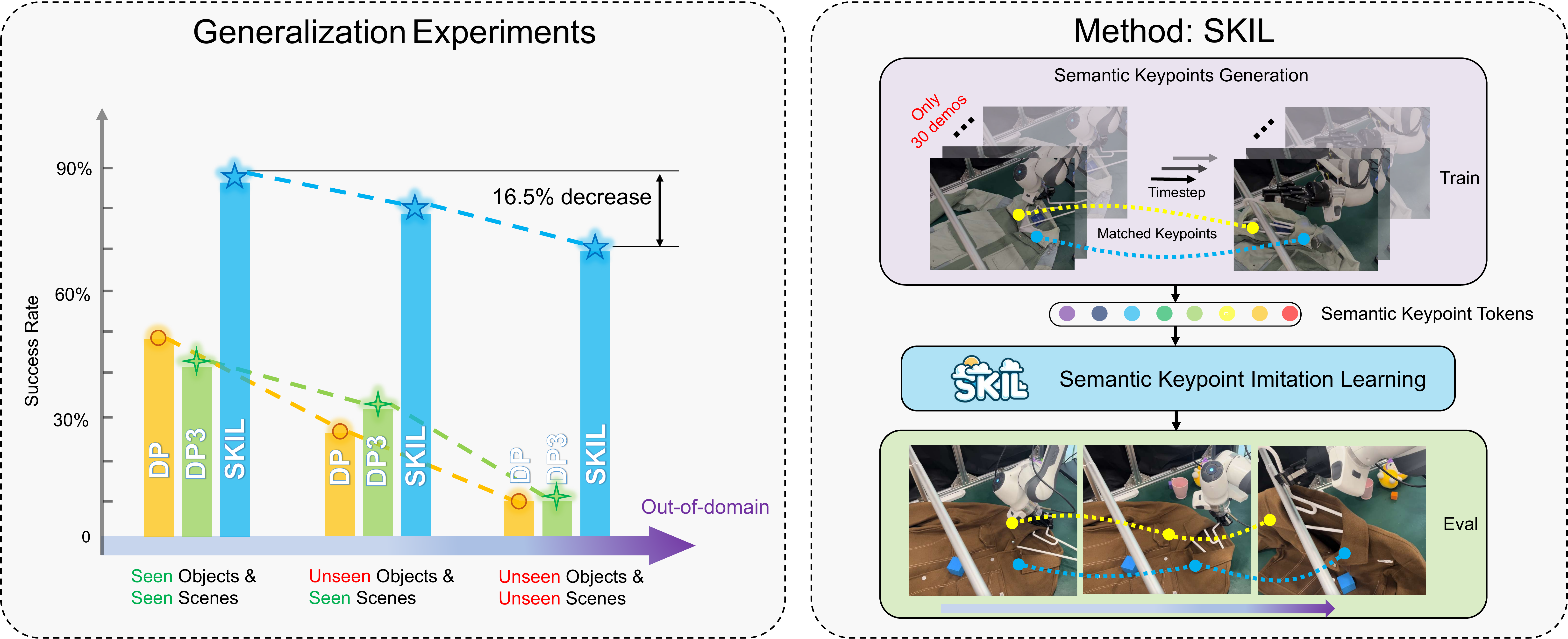}
    \captionof{figure}{Given limited robot demonstrations, our method, \textbf{S}emantic \textbf{K}eypoint \textbf{I}mitation \textbf{L}earning (SKIL), can achieve superior performance for generalizable and long-horizon manipulation tasks, such as hanging a cloth on a rack. When encountering unseen objects and scenes, SKIL outperforms baselines by a large margin in four short-horizon tasks.}
\end{center}%
}]

\IEEEpeerreviewmaketitle


\begin{abstract}

Real-world tasks such as garment manipulation and table rearrangement demand robots to perform generalizable, highly precise, and long-horizon actions. Although imitation learning has proven to be an effective approach for teaching robots new skills, large amounts of expert demonstration data are still indispensible for these complex tasks, resulting in high sample complexity and costly data collection. 
To address this, we propose Semantic Keypoint Imitation Learning (SKIL), a framework which automatically obtain semantic keypoints with help of vision foundation models, and forms the descriptor of semantic keypoints that enables effecient imitation learning of complex robotic tasks with significantly lower sample complexity. In real world experiments, SKIL doubles the performance of baseline methods in tasks such as picking a cup or mouse, while demonstrating exceptional robustness to variations in objects, environmental changes, and distractors. For long-horizon tasks like hanging a towel on a rack where previous methods fail completely, SKIL achieves a mean success rate of 70\% with as few as 30 demonstrations. 
Furthermore, SKIL naturally supports cross-embodiment learning due to its semantic keypoints abstraction, our experiments demonstrate that even human videos bring considerable improvement to the learning performance. All these results demonstrate the great success of SKIL in achieving data-efficint generalizable robotic learning.
Visualizations and code are available at: \href{https://skil-robotics.github.io/SKIL-robotics/}{https://skil-robotics.github.io/SKIL-robotics/}.

\end{abstract}    
\section{Introduction} \label{sec:intro}
End-to-end policy learning has gained significant attention in training robotic systems \cite{levine2016end,chen2023visual,zare2024survey}. Imitation learning, in particular, has been instrumental in enhancing the efficiency of end-to-end training by enabling robots to learn directly from expert demonstrations via supervised learning \cite{peng2020learning,radosavovic2021state,zhao2023learning}. 
While current methods have shown success in various robotic manipulation scenarios, many real-world tasks, such as garment manipulation and table rearrangement, require policies that are generalizable and capable of highly precise or long-horizon actions \cite{fu2024mobile,zhao2024aloha,fu2024humanplus}. 

We take the household task of hanging clothes as an illustrative example. Hanging clothes on a rack involves multiple stages, such as selecting a hanger, precisely inserting the garment onto the hanger, and subsequently placing it on the rack. This task also requires generalization across different types of clothing and varying positions. Consequently, the task inherently demands a large number of demonstrations, resulting in high sample complexity. However, data collection for such a complex task is both time-consuming and costly. A recent work \cite{zhao2024aloha} has demonstrated similar skills through nearly 10 thousand robot demonstrations.
To address this challenge, prior approaches have focused on advancing data collection methods \cite{zhao2024aloha, ha2024umi, chi2024universal, gao2024efficient, black2024pi_0, lin2024data} and developing advanced representation techniques \cite{nair2022r3m, ke20243d, zhang2023universal, wang2024rise, liu2024rdt, black2024pi_0}. For instance, dataset-based methods emphasize the importance of constructing diverse datasets \cite{zhao2024aloha, gao2024efficient, lin2024data}. A recent study finds that collecting data in a wide range of environments, each with unique manipulation objects and accompanying demonstrations, significantly enhances generalization capabilities \cite{lin2024data}. However, even with these strategies, achieving zero-shot generalization to novel objects and environments may require a large amount of demonstrations for a single task \cite{lin2024data}.
In parallel, research into novel representations—such as pre-trained vision models \cite{nair2022r3m, di2024dinobot, kim2024openvla, black2024pi_0}, 3D visual representations \cite{gervet2023act3d, ze20243d, wang2024rise}, object-centric representations \cite{shi2024composing, qian2024task}, and semantic-geometric features \cite{ke20243d, zhang2023universal, wang2024gendp}—has aimed to overcome these limitations. These advanced representations improve sample efficiency, particularly by exhibiting spatial generalization \cite{ze20243d, wang2024rise}. Despite this progress, these methods still tend to overfit to the seen objects and scenes during training, struggling to handle unseen objects and environments.
Given these challenges, a critical question arises: \textit{How can we reduce sample complexity to enable robots to learn data-efficient and generalizable manipulation tasks?}

In this paper, we propose Semantic Keypoints Imitation Learning (SKIL), an imitation learning framework that leverages a vision foundation model to identify semantic keypoints as observations. This sparse representation reduces the problem's dimensionality, thereby achieving a lower sample complexity. By matching consistent keypoints between training and testing objects, SKIL utilizes their associated features and spatial information as conditional inputs to a diffusion-based action head, which outputs the robot's actions. 
Additionally, SKIL inherently facilitates cross-embodiment learning through its abstraction of semantic keypoints. 
Our key contributions are summarized as follows:
\begin{enumerate} 
    \item[\textit{1)}] We propose the Semantic Keypoint Imitation Learning (SKIL) framework, which automatically obtains the semantic keypoints through a vision foundation model, and forms the descriptor of semantic keypoints for downstream policy learning.
    \item[\textit{2)}] SKIL offers a series of advantages. First, the sparsity of semantic keypoint representations enables data-efficient learning. Second, the proposed descriptor of semantic keypoints enhances the policy's robustness. Third, such semantic representations enables effective learning from cross-embodiment human and robot videos.
    \item[\textit{3)}] SKIL shows a remarkable improvement over previous methods in 6 real-world tasks, by achieving a success rate of 72.8\% during testing, offering a 146\% increase compared to baselines. SKIL can perform long-horizon tasks such as \textit{hanging a towel or cloth on a rack}, with as few as 30 demonstrations, where previous methods fail completely.
\end{enumerate}


\section{Related Work}
\subsection{Imitation Learning}
Imitation learning from expert demonstrations has always been an effective approach for teaching robots skills \cite{peng2020learning,radosavovic2021state,zhao2023learning,fu2024humanplus}, in which behavior cloning (BC) serves as a most basic and straight-forward algorithm by directly taking expert actions as supervision labels \cite{torabi2018behavioral,florence2022implicit}. Considering the challenges of obtaining accurate states while implementing in real-world environments, the most intuitive yet simple idea, end-to-end mapping from images to actions, has become one of the most popular choices of researchers in recent years \cite{xu2017end,zhao2023learning,chi2023diffusion}.
For example, the ACT algorithm employs a transformer architecture to produce action tokens from encoded image tokens, and achieves accurate closed-loop control \cite{zhao2023learning,zhao2024aloha,fu2024mobile}. Diffusion Policy, on the other hand, leverages the diffusion process to model a conditional action distribution, therefore achieving multimodal behavior learning ability and stabler training \cite{chi2023diffusion,chi2024universal,ha2024umi}.


Given the advantages of Diffusion Policy \cite{chi2023diffusion}, recent research has focused on improving its representation capabilities. 
Some recent methods explore how to fuse information from 3D visual scenes, language instructions, and proprioception \cite{gervet2023act3d,shridhar2023perceiver,zhang2023universal,zhang2024leveraging,ke20243d}. However, these approaches typically predict keyframes rather than continuous actions (e.g., Peract \cite{shridhar2023perceiver}, Act3D \cite{gervet2023act3d}, 3D Diffusor actor \cite{ke20243d}), which makes them less effective at completing complex tasks.
Other methods such as DP3 \cite{ze20243d}, RISE \cite{wang2024rise} and EquiBot \cite{yang2024equibot} utilize 3D perception as observations and output the sequence of continuous actions. However, as demonstrated in our experiments, these methods severely lacks real-world generalization abilities with limited demonstrations.
Furthermore, GenDP \cite{wang2024gendp} computes dense semantic fields via cosine similarity with 2D reference features, and achieves category-level manipulation by taking semantic fields as input.
However, semantic fields contain too much redundancy information, which harms the learning efficiency. 
In contrast, our method leverages semantic keypoints to construct a sparse representation, which, when conditioned on the policy, reduces the need of amount of demonstrations.

\subsection{Keypoint-based Imitation Learning}
Extracting point motions from visual images serves as a general feature representation method. Due to its inherent sparsity, this approach has proven to be data-efficient in robotic manipulation \cite{weng2022fabricflownet,eisner2022flowbot3d,seita2023toolflownet}. 
Early works typically required supervised training on large datasets from simulators or real world, to learn motion of points on related objects\cite{zhang2023flowbot++,eisner2022flowbot3d,vecerik2024robotap}.
Recent approaches, such as ATM \cite{wen2023any}, Track2Act \cite{bharadhwaj2024track2act}, GeneralFlow \cite{yuan2024general} and Im2Flow2Act \cite{xu2024flow}, utilize an off-the-shelf tracker (e.g., Cotracker \cite{karaev2025cotracker}) to observe the motion of points. 
These models support human-to-robot transfer by leveraging these point motion trajectories during policy training.
However, despite these advances, observing accurate point motions remains challenging, as it struggles to generalize to unseen objects.
Keypoint representation dramatically reduces the dimensionality of the state, thereby achieving high efficiency in robot navigation and manipulation \cite{mur2015orb,forster2014svo,florence2019self,di2024dinobot}. Learning-based methods for keypoint extraction require large datasets and self-supervised training to generalize across object categories \cite{florence2019self,manuelli2019kpam, xue2023useek}. Recent advances in vision models, such as DINOv2 \cite{oquab2023dinov2} and DiFT \cite{tang2023emergent}, allow the use of pre-trained models to extract semantic correspondence. DINOBot \cite{di2024dinobot} and Robo-ABC \cite{ju2025robo} can retrieve visually similar objects from human demonstrations and align the robot's end-effector with new objects. However, this approach lacks feedback loops, limiting its use to offline planning.

Some recent works showed success in learning short-horizon tasks rapidly.
ReKep \cite{huang2024rekep} utilizes DINOv2 \cite{oquab2023dinov2} for getting keypoint proposals and GPT-4 \cite{hurst2024gpt} for building relational constraints of keypoints, and then applies an optimization solver to generate robot trajectories. KALM \cite{fang2024keypoint} leverages Segment Anything (SAM) \cite{kirillov2023segment} and large language models (LLMs) to automatically generate task-relevant, consistent keypoints across instances. KAT \cite{di2024keypoint} employs in-context learning (ICL) with LLMs, requiring only 5–10 demonstrations to teach the robot new skills, and their subsequent work, Instant Policy \cite{vosylius2024instant}, extends ICL to a graph generation problem.
However, these methods struggle to achieve precise or long-horizon motion planning due to the inaccuracy and latency of LLMs. In contrast, our method utilizes semantic keypoints as observations and applies a diffusion action head for real-time imitation learning with continuous actions.
A very recent work \cite{levy2024p3} also uses semantic keypoints as observations of an imitation learning algorithm. However, they rely on off-the-shelf tracking models \cite{karaev2025cotracker} to derive keypoints' positions, which restricts their application to long-horizon tasks. In contrast, our method SKIL enables learning long-horizon tasks such as hanging a towel or cloth on a hanger, with only 30 demonstrations.
\section{Method}
Our proposed method, SKIL, comprises two primary modules, which is shown in Figure \ref{fig:method}. 
Based on RGBD input frames, the \textit{Semantic Keypoints Description Module} first obtains the semantic keypoints and computes the descriptor of keypoints (Section \ref{sec:descriptor_kp}). The \textit{Policy Module} then uses a transformer encoder to fuse the information of keypoints descriptor, and finally applies a diffusion action head to output robot actions (Section \ref{sec:policy_module}).
We also introduce an extra cross-embodiment learning version of SKIL in Section \ref{sec:cross-embody}.


\subsection{Key Insight}
Previous perception modules often tend to overfit specific training objects and scenes, struggling to handle objects with varying colors, textures, and geometries. However, practical manipulation tasks rely little on these detailed properties. For instance, when picking up a cup, a smart agent should focus mainly on the position of the handle instead of its color or shape. Similarly, when folding clothes, the positions of the collar and sleeves matter the most. In this context, sparse semantic keypoints, such as the handle of a cup or the collar and sleeves of a shirt, serve as the most critical task-relevant information. 
These keypoints remain highly consistent across different objects or scenes, enabling them to address the overfitting challenge. Furthermore, this simplified formulation can significantly reduce the need of extensive demonstrations, reaching a much higher sample efficiency.

Recently, vision foundation models have showed remarkable success across various downstream tasks, particularly excelling in semantic correspondence detection \cite{oquab2023dinov2, kirillov2023segment, tang2023emergent, ranzinger2024radio}. This success motivates us to leverage vision foundation models to extract keypoints with semantic correspondence, as introduced later in the following sections.



\subsection{Semantic Keypoints Description Module}
\label{sec:descriptor_kp}
\begin{figure*}
    \centering
  \includegraphics[width=\linewidth]{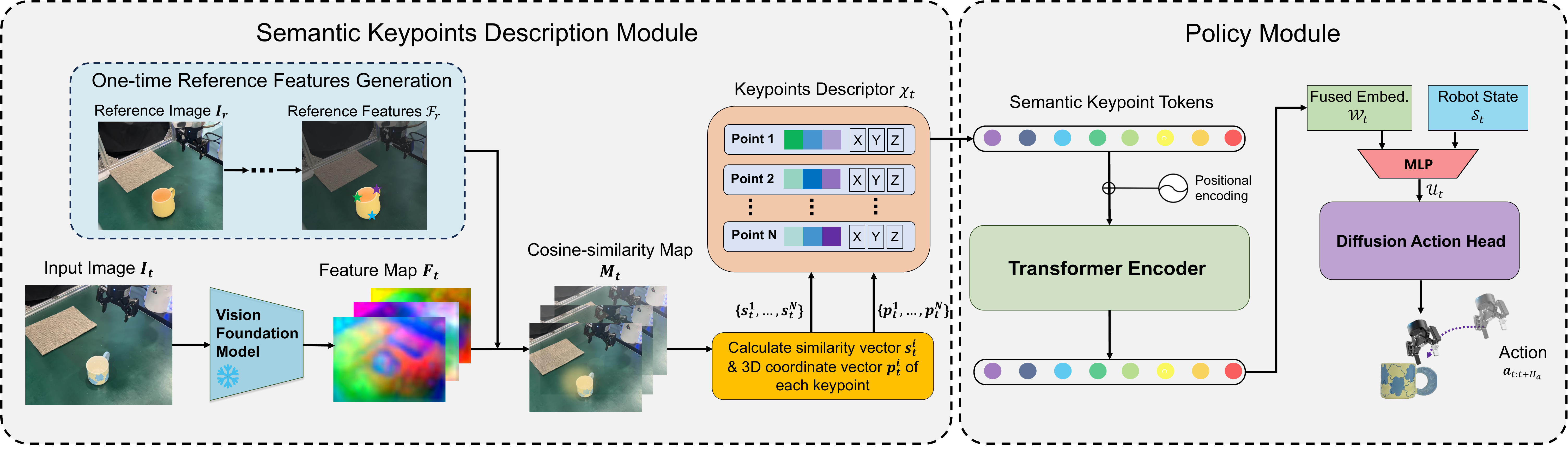}
  \caption{Overview of our framework SKIL, including \textit{Semantic Keypoints Description Module} and \textit{Policy Module}. The first module computes descriptors for the semantic keypoints. Then, we apply a transformer encoder to obtain the fused embedding of the keypoints. Conditioned on the fused embedding and robot state, a diffusion action head outputs the final action sequence.}
  \label{fig:method}
\end{figure*}

In this module, we first obtain the reference features of the task, with the help of a vision foundation model. 
Based on the reference features, we build the cosine-similarity map of the current frame.
Finally, we calculate the descriptor of semantic keypoints from the cosine-similarity map and the original depth image. The process can be seen in Figure \ref{fig:method}.

\begin{figure}[t]
    \centering
    \includegraphics[width=\linewidth]{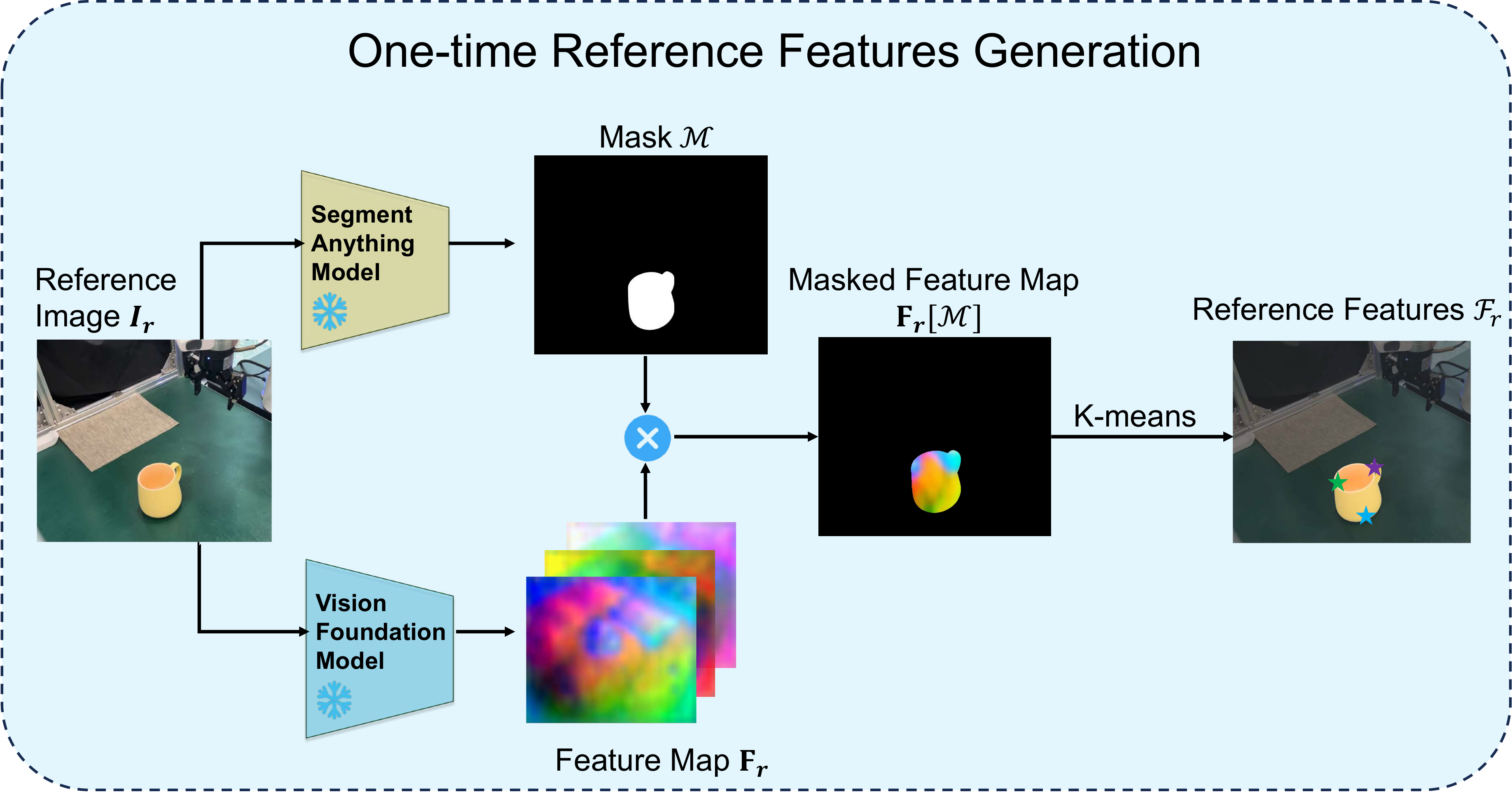}
    \caption{Process of generating reference features, given a single reference image of the specific task: (1) Apply SAM \cite{kirillov2023segment} and Vision Foundation Model to obtain the mask $\mathcal{M}$ and the feature map $\mathbf{F}_r$ individaully; (2) Cluster the masked features $\mathbf{F}_r[\mathcal{M}]$ to obtain the reference features $\mathcal{F}_r$ using K-means.}
    \label{fig:reference_buffer}
\end{figure}

\textbf{One-time Reference Features Generation.} 
For each task, we only require one single image of the task scene to automatically detect the reference keypoints and features, which are then used throughout the entire training and evaluation process. We illustrate this one-time reference features generation process in Figure \ref{fig:reference_buffer}. Given an RGB reference image $\mathcal{I}_r \in \mathbb{R}^{H \times W \times 3}$, we first extract patch-wise features using a vision foundation model (e.g., DiFT \cite{tang2023emergent} and RADIO \cite{ranzinger2024radio}) and apply bilinear interpolation to upsample the features to the original image size, $\mathbf{F}_r \in \mathbb{R}^{H \times W \times D}$. Meanwhile, we use Segment Anything Model (SAM) \cite{kirillov2023segment} to generate a mask $\mathcal{M}$ of all relevant objects. We then combine these two results to get the masked feature map $\mathbf{F}_r [\mathcal{M}]$, which contains $|\mathcal{M}|$ non-zero feature vectors of dimension $D$. Finally, we apply K-means to cluster these feature vectors into $N$ clusters, with center pixel positions $\left\{(h^i_r,w^i_r)\in\mathcal{M}\mid 0<i\leq N\right\}$. These cluster centers forms the reference keypoints, and their correspoinding features form the set of reference features, which is $ \mathcal{F}_r =\left\{ \mathcal{F}^i_r = \mathbf{F}_r[h^i_r,w^i_r]\in \mathbb{R}^{D}\mid 0 < i \leq N \right\}$.

Note that $N$ is a manually set hyperparameter, and K-means could be replaced by other keypoint proposal strategies. See Section \ref{sec:ablation} for more detailed discussions. 

\textbf{Cosine-similarity Map Generation.} As shown in Figure \ref{fig:method}, during the training and inference phases, the input image $\mathcal{I}_t \in \mathbb{R}^{H \times W \times 3}$ at current timestep is processed by the same vision foundation model to obtain the feature map at the original image size, $\mathbf{F}_t \in \mathbb{R}^{H \times W \times D}$. 
We compute the cosine-similarity map between  $\mathbf{F}_t$ and the reference features $\mathcal{F}_r$, 
\begin{equation}
    \label{eq:cos_map}
    \mathbf{M}_{t}= \text{cosine\_sim} (\mathbf{F}_t,  \mathcal{F}_r),
\end{equation}
where $\mathbf{M}_{t} \in \mathbb{R}^{H \times W \times N}$, whose $i$-th channel (denoted as $\mathbf{M}^i_{t}$ later) among the $N$ channels represents the cosine-similarity map between the current frame $\mathcal{I}_t$ and the $i$-th reference feature.

\textbf{Keypoints Descriptor Calculation.} 
According to the similarity map $\mathbf{M}_{t}$, we can obtain the pixel coordinate $(h_t^i, w_t^i)$ of each matched semantic keypoint, denoted as follows:
\begin{equation}
    \left(h_t^i, w_t^i\right) = 
        \mathop{\arg\max}\limits_{(h,w)} (\mathbf{M}^i_{t}[h,w]) \ ,0<i\leq N.
\end{equation}
The pixel coordinate of each keypoint can serve as the intermediate representation in some flow-based polices, such as ATM \cite{wen2023any} and Track2Act \cite{bharadhwaj2024track2act}. However, this representation is lacking for semantic and spatial description of keypoints, harming the downstream policy learning. 

Therefore, we compute a descriptor for each matched keypoint, consisting of a similarity vector and a 3D coordinate vector. The similarity vector represents the cosine-similarities between the matched keypoint and all reference keypoints. The vector can identify the matched keypoint by its maximum value, and the magnitude of this value represents the confidence of this matching. Since the similarity map $\mathbf{M}_{t}$ stores the cosine-similarities between all pixels of the input image and reference keypoints, the similarity vector can be defined as $\mathbf{s}^i_t \in \mathbb{R}^{N}$ : 
\begin{equation}
\label{eq:descriptor-kp}
\begin{split}
\mathbf{s}^i_t &= \mathbf{M}_{t} \left[h_t^i, w_t^i, \cdot\ \right] \ ,0<i\leq N.
\end{split}
\end{equation}
Based on the pointcloud derived from the depth image, we obtain the 3D coordinate vector of each matched keypoint, defined as $\mathbf{p}^i_t \in \mathbb{R}^3$. 
Overall, the descriptor of each matched keypoint can be denoted by 
\begin{equation}
\label{eq:descriptor-chi}
\begin{split}\mathbf{\chi}^i_t = [\mathbf{s}^i_t, \mathbf{p}^i_t] \ ,0<i\leq N,
\end{split}
\end{equation}
which is later fed into the next \textit{Policy Module}.

\subsection{Policy Module} 
\label{sec:policy_module}
\textbf{Transformer Encoder.}~\label{sec:sk_encoder}
We first tokenize each descriptor $\chi^i_t$ into tokens of each keypoint. Specifically, each descriptor is first embedded into a $d$-dimensional latent space with positional encoding. 
As shown in Figure \ref{fig:method}, a transformer encoder processes all tokens and we compute the mean of all output tokens to obtain the fused embedding of keypoints $\mathcal{W}_t$. We define this whole process as
\begin{equation}
    \mathcal{W}_t = \text{Encoder}\left(\mathbf{\chi}^1_t,\mathbf{\chi}^2_t,...,\mathbf{\chi}^N_t\right)
\end{equation}
where $t$ denotes the timestep, $N$ denotes the number of keypoints. 
Note that we choose mean of tokens \cite{radford2021learning} instead of a [CLS] token, for its slightly better performance in our experiments.

\textbf{Diffusion Action Head.}~\label{sec:policy}
Based on the aforementioned encoder, we obtain the fused embedding $\mathcal{W}_t$ of the keypoints. 
We concatenate $\mathcal{W}_t$ with the robot state $\mathcal{S}_t$ (including joint positions, end-effector position and orientation, gripper state, etc) and use a multi-layer perceptron (MLP) to fuse them into a compact representation 
\begin{equation}
    \mathcal{U}_t = \text{MLP}\left(\mathcal{S}_t, \mathcal{W}_t\right),
\end{equation}
as shown in Figure \ref{fig:method}. 

Conditioned on the compact representation $\mathcal{U}_t$, a diffusion action head outputs the robot action. Following Diffusion Policy (DP) \cite{chi2023diffusion}, we use a CNN-based U-Net as the noise prediction network. Detailed formulations are provided in Appendix \ref{app:diff_head}.
To improve temporal consistency, we predict an action chunk in a single step, $\mathbf{a}_{t:t+H_a}:= (\mathbf{a}_t, \dots, \mathbf{a}_{t+H_a-1})$, where $H_a$ denotes the chunk size. 
For real-time inference, we utilize DDIM \cite{song2020denoising}, a diffusion model sampling accelerator, to reduce the number of diffusion denoising steps.

\textbf{Action Ensemble.} 
Existing vision foundation models occasionally produce mismatching of keypoints, which causes motion jitter. 
To address this issue, we employ an ensemble approach for action planning. Specifically, during training, we randomly dropout 20\% of the semantic keypoint tokens of each frame, before sending them to the transformer encoder. 
During testing, we repeat the action inferring process (with this random dropout) 20 times, and get the median of all output actions to be the finally executed action. 
(Note that this repetition can be done parallelly across the batch dimension, so that introduces almost no extra latency.)
This ensemble strategy ensures smoother and more reliable action execution.





\subsection{Cross-embodiment Learning}
\label{sec:cross-embody}
In this section we define an extra cross-embodiment learning version of SKIL. Our motivation is that semantic keypoints abstraction avoids incorporating embodiment information, therefore enables the use of diverse data source (including human videos). Inspired by ATM \cite{wen2023any}, a cross-embodiment learning framework, we view the trajectory prediction of keypoints as an intermediate task. The predicted trajectories serve as effective guidance for learning policies.
We name this cross-embodiment version \textbf{SKIL-H}, which involves 2 modules: 
\begin{enumerate}
    \item \textit{Trajectory Prediction Module}:
    \begin{itemize}[leftmargin=5pt]
        \item predicts future keypoint positions from pure video data, 
        \item trained with both robot and human demonstrations;
    \end{itemize}
    \item \textit{Trajectory-to-Action Module}:
    \begin{itemize}[leftmargin=5pt]
        \item maps the predicted trajectories into robot actions,
        \item trained with only robot demonstrations.
    \end{itemize}
\end{enumerate}

\begin{figure}[ht]
    \centering
    \includegraphics[width=\linewidth]{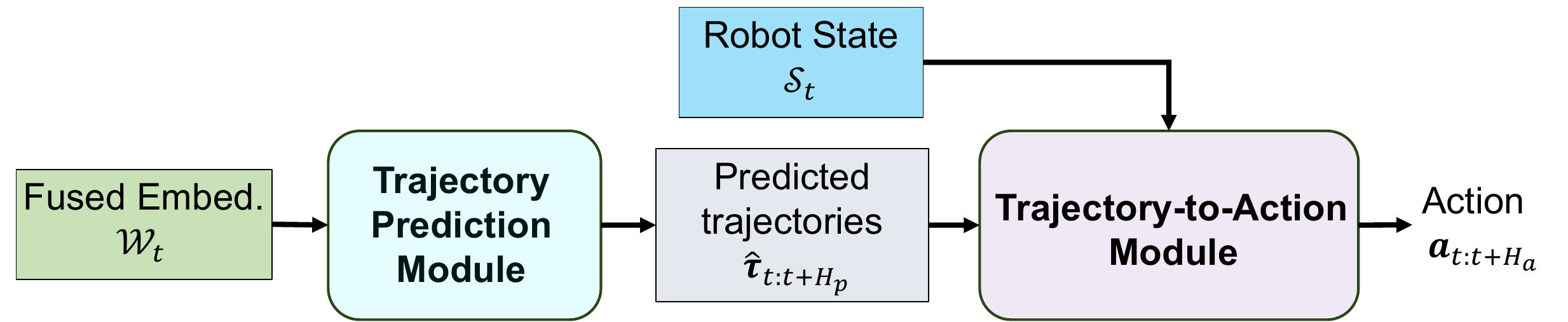}
    \caption{Architecture of SKIL-H, comprising the \textit{Trajectory Prediction Module} and the \textit{Trajectory-to-Action Module}. The first module predicts the trajectories $\mathbf{\hat\tau}_{t:t+H_p}$ of matched keypoints based on the fused embedding $\mathcal{W}_t$. The second module takes the predicted trajectories $\mathbf{\hat\tau}_{t:t+H_p}$ and the robot state $\mathcal{S}_t$ as inputs, and outputs the robot action sequence $\mathbf{a}_{t:t+H_a}$.}
    \label{fig:pred_skil_model}
\end{figure}

As illustrated in Figure \ref{fig:pred_skil_model}, at timestep $t$, the \textit{Trajectory Prediction Module} of SKIL-H takes the fused embedding $\mathcal{W}_t$ (produced by original SKIL) as input, and predictes the future keypoint trajectories as 
\begin{equation}
    \mathbf{\hat \tau}_{t:t+H_p} = \left\{\mathbf{\hat p}^i_q \mid t < q \le t + H_p,\  0 < i \leq N\right\},
\end{equation}
in which $\mathbf{\hat p}^i_q$ denotes the predicted 3D position of $i$-th matched keypoint at future timestep $q$, \(N\) is the number of predicted keypoints and \(H_p\) is the prediction horizon.
We employ a diffusion model to build the \textit{Trajectory Prediction Module}. 
The training labels of the model are obtained with the help of an off-the-shelf tracking model (e.g., CoTracker \cite{karaev2025cotracker}). 
Specifically, we obtain the 2D flow of the matched keypoints from videos using the tracking model and project them back to 3D real trajectories $\mathbf{\tau}_{t:t+H_p}$, as the training labels. 

The next \textit{Trajectory-to-Action Module} of SKIL-H takes the predicted trajectories $\mathbf{\hat\tau}_{t:t+H_p}$ and the robot state $\mathcal{S}_t$ as input, and process them with a transformer encoder followed by a diffusion action head to output the final robot action $\mathbf{a}_{t:t+H_a}$. This module functions similarly as the origin \textit{Policy Module} of SKIL (See section \ref{sec:policy}), but with different input format and encoder architecture.
All other settings including the training loss remain the same. 



\begin{figure*}[ht]
    \centering
    \includegraphics[width=0.90\linewidth]{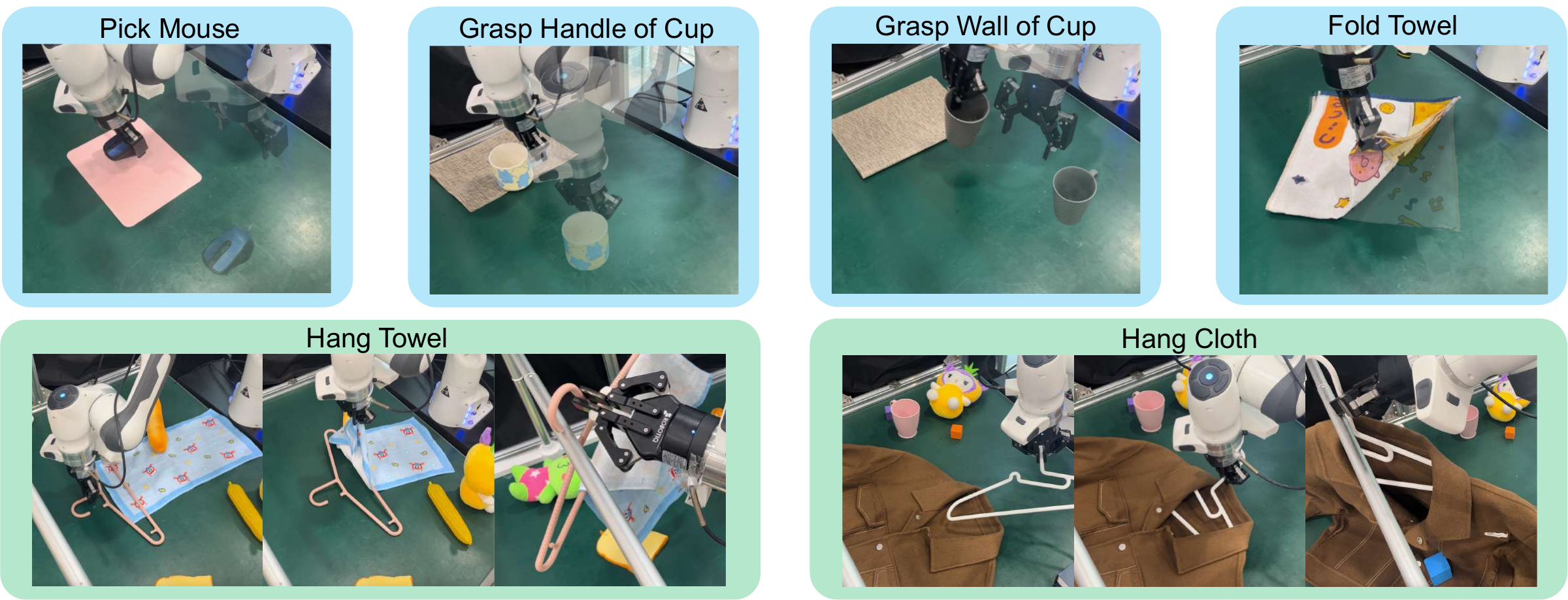}
    \caption{Overview of our 6 real-world tasks, including the top 4 short-horizon ones and the bottom 2 long horizon ones. }
    \label{fig:real_exp}
\end{figure*}

\begin{table*}[ht]
  \centering
  \caption{Realworld results, measured by evaluation phase success rates on the unseen testing objects. SKIL outperforms baseline methods by a large margin on either training or testing objects. (All objects shown in Figures \ref{fig:objects-1} and \ref{fig:objects-2}) Testing objects are unseen but belong to the same categories. For each object, we conduct five trials with random initialization. (The baselines with point clouds input perform poorly on \textit{Grasp Wall of Cup}. We discuss a possible reason for this in Appendix \ref{app:cupwall_analysis}.)}
  \begin{tabular}{@{}cccccccccccccccc@{}}
    \toprule
    \multicolumn{1}{l|}{Method/Task} & \multicolumn{2}{c|}{\textit{Pick Mouse}} & \multicolumn{2}{c|}{\textit{Grasp Handle of Cup}} & \multicolumn{2}{c|}{\textit{Grasp Wall of Cup}} & \multicolumn{2}{c|}{\textit{Fold Towel}} & \multicolumn{2}{c|}{\textit{Hang Towel}} & \multicolumn{2}{c|}{\textit{Hang Cloth}} & \multicolumn{2}{c}{Average}\\
    \multicolumn{1}{l|}{}& Train  & \multicolumn{1}{c|}{Test} & Train & \multicolumn{1}{c|}{Test} & Train  & \multicolumn{1}{c|}{Test} & Train & \multicolumn{1}{c|}{Test} & Train   & \multicolumn{1}{c|}{Test} & Train & \multicolumn{1}{c|}{Test} & Train & \multicolumn{1}{c}{Test}\\
    \midrule
    DP & 40\% & 4\% & 30\% & 36\% & 70\% & 36\% & 60\% & 50\% & 0\% & 0\% & 20\% & 12\% & 36.7\% & 23.0\%\\
    DP3 & 20\% & 18\% & 50\% & 32\% & 30\% & 18\% & 60\% & 58\% & 0\% & 0\% & 30\% & 24\% & 31.6\% & 25.0\% \\
    RISE & 10\% & 14\% & 40\% & 34\% & 10\% & 12\% & 50\% & 32\% & 0\% & 0\% & 20\% & 16\% & 21.7\% & 18.0\% \\
    GenDP-S & 40\% & 32\% & 50\% & 32\% & 40\% & 30\% & 60\% & 58\% & 0\% & 0\% & 30\% & 28\% & 36.7\% & 30.0\%\\
    SKIL (Ours) & \textbf{90\%} & \textbf{72\%} & \textbf{90\%} & \textbf{94\%} & \textbf{90\%} & \textbf{80\%} & \textbf{90\%} & \textbf{72\%} & \textbf{70\%} & \textbf{72\%} & \textbf{60\%} & \textbf{47\%} & \textbf{81.7\%} & \textbf{72.8\%} \\
    \bottomrule
  \end{tabular}
  \label{tab:real-results}
\end{table*}


\section{Experiment Setup}
We introduce the experiment setup in this section, including the task definitions, data collection \& evaluation settings, and baselines to be compared with. 
\subsection{Task Definitions}
We use a Franka robot arm equipped with a Robotiq gripper to perform six real-world tasks, including the first four short-horizon tasks and the last two long-horizon ones. A brief overview of these tasks is listed below: (Visualizations are provided in Figure \ref{fig:real_exp}.)
\begin{enumerate}
    \item \textbf{Pick Mouse}: The gripper grasps a mouse from the workspace and places it on the mouse mat.
    \item \textbf{Grasp Handle of Cup}: The gripper grasps the cup's handle and places the cup on the right side of the table.
    \item \textbf{Grasp Wall of Cup}: Instead of the handle, the gripper grasps the wall of the cup and places it on the right side of the table.
    \item \textbf{Fold Towel}: The gripper grasps the left corner of the towel and lifts it toward the right corner.
    \item \textbf{Hang Towel}: This multi-step task involves grasping a hanger from the table, placing it near the towel, pinching the towel's top edge to fold it through the hanger, and hanging the hanger on a rack. Visualization is provided in Figure \ref{fig:long-horizon-towel} in Appendix \ref{app:long-horizon}.
    \item \textbf{Hang Cloth}: This task involves grasping a hanger from the table, precisely inserting it into the cloth collar, rotating the hanger, and hanging the cloth on the rack. Visualization is provided in Figure \ref{fig:long-horizon-cloth} in Appendix \ref{app:long-horizon}.
\end{enumerate}
The object poses and the joint positions of the Franka arm are randomly initialized throughout data collection and evaluation.
See more details in Appendix \ref{app:randomization}. 


Besides, we also conduct experiments on several simulation tasks. We select ten tasks from the MetaWorld \cite{yu2020meta} and DexArt \cite{bao2023dexart} benchmarks. More details about the simulation tasks can be found in Appendix \ref{app:simenv}.

\subsection{Data Collection \& Evaluation}
Real-world expert demonstrations are collected through human teleoperation, following the collection process of Droid dataset \cite{khazatsky2024droid}. The hardware setup is described in Appendix \ref{app:hardware}, where a Franka arm with a Robotiq gripper is teleoperated using a Meta Quest controller \cite{khazatsky2024droid}. 
We collect 20 demonstrations for short-horizon tasks and 30 demonstrations for long-horizon tasks respectively.
For all six tasks, we use 2 objects for training data collection, and we use 10 objects for the first four short-horizon tasks and 3$\sim$5 objects for the last two long-horizon ones during evaluation.

The action space contains the end-effector pose and gripper state, while observations include RGB images and corresponding depth images captured by a fixed third-view Zed2 camera, as shown in Figure \ref{fig:hardware} in Appendix \ref{app:hardware}.

As for the simpler simulation tasks, we collect 10 expert demonstrations for the chosen MetaWorld \cite{yu2020meta} and DexArt \cite{bao2023dexart} tasks. More detailed settings can be found in Appendix \ref{app:simenv}.

For all real-world and simulation tasks, we measure the performance of a specific method by its average success rate (on unseen testing objects for most tasks) in the evaluation phase. 

\subsection{Baselines}
We compare SKIL with state-of-the-art imitation learning algorithms. \textbf{Diffusion Policy (DP)} \cite{chi2023diffusion} models the action distribution using a diffusion model and leverages RGB observations as conditions in the diffusion model. \textbf{DP3} \cite{ze20243d} utilizes a similar diffusion architecture to Diffusion Policy and introduces a compact 3D representation instead of 2D images by employing an efficient MLP encoder. \textbf{RISE} \cite{wang2024rise} uses 3D point clouds to predict robot actions by first processing the data with a shallow 3D encoder and then mapping it to actions using a transformer. \textbf{GenDP-S}: GenDP \cite{wang2024gendp} generates 3D descriptor fields from multi-view RGBD data, computes semantic fields via cosine similarity with 2D reference features, and uses PointNet++ and a diffusion model to predict robot actions.
Note that we name our implementation GenDP-S because we build 3D descriptor fields from single view. More implementation details of SKIL and all these baselines can be found in Appendix \ref{app:train-detail} and \ref{app:baselines} respectively. 

\section{Results \& Analysis}\label{sec:experiment}
In this section, we present SKIL's performance along with its comparison result with baseline methods, from which we can prove the strong generalization ability and the excelling data efficiency of SKIL. We also demonstrate the performance of SKIL-H, showing its cross-embodiment learning ability. Finally, we present ablation studies to assess our choices on each of SKIL's components.

\begin{figure*}[ht]
  \centering
  \includegraphics[width=0.94\linewidth]{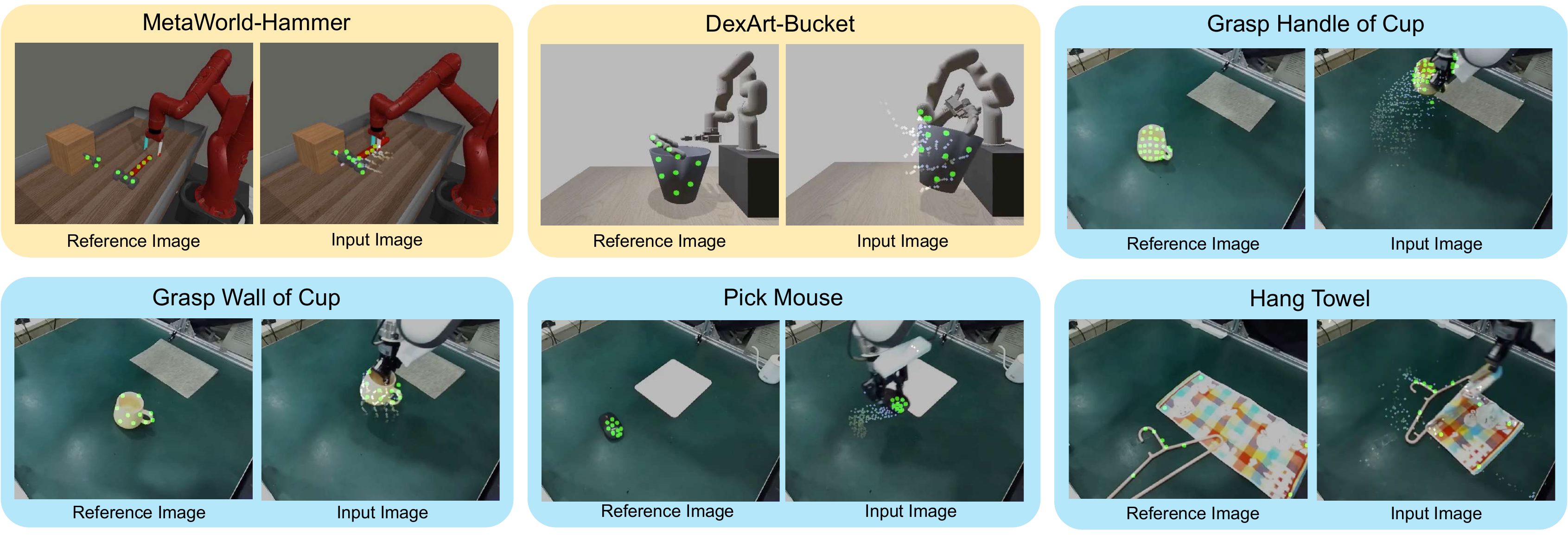}
  \caption{Movement of semantic keypoints in SKIL. Green points represent the keypoints in the current frame, and the white trajectories show their movements in previous timesteps. Most keypoints maintain temporal consistency. Although some keypoints are occluded, the mismatched points remain close to the relevant object.}
  \label{fig:keypoint-movement}
\end{figure*}

\subsection{Performance \& Comparison}

Table \ref{tab:real-results} presents the main results on real-world tasks. SKIL significantly outperforms several strong baselines across all tasks, by achieving a mean success rate of 72.8\% under unseen objects, comparing to the highest success rate of 30\% achieved by baselines. Figure \ref{fig:real_exp} presents snapshots of the real-world experiments.
SKIL also reaches the best performance across the baselines on simulation tasks. Detailed results and analysis of simulation results are provided in Appendix \ref{app:sim-results}.

For an intuitive glance of SKIL's keypoint-based representation, Figure \ref{fig:keypoint-movement} illustrates the moving trajectories of the matched semantic keypoints on several tasks. By comparing with the keypoints in the reference images, we observe that most of the matched semantic keypoints in the input images maintain high temporal consistency. Even when some keypoints are occluded, the mismatched points remain close to the correct position. Additional visualizations of keypoint trajectories on different objects can be found in Appendix \ref{app:keypoint-movement}.

In the following, we analyze the generalization ability and data efficiency of SKIL with specific examples. 

\subsubsection{\textbf{Generalization}}
We categorize generalization into three dimensions: \textit{Spatial generalization}, \textit{Object generalization} and \textit{Environment generalization}.

\paragraph{\textbf{Spatial generalization}} We can see that the baselines achieve inferior performance on the \textit{Pick Mouse} task. These methods often fail to grasp the mouse that is near the corner of the workspace. In contrast, SKIL is able to handle most of the workspace, as illustrated in Figure \ref{fig:mouse-corner} in Appendix \ref{app:spatial_gen}. Meanwhile, SKIL can pick up the towel's corner more precisely than the baselines, as shown in Figure \ref{fig:towel-failure} in Appendix \ref{app:spatial_gen}. 
The improvement is primarily due to the semantic keypoints located on relevant objects, which helps the policy better understand the pose of the objects.

\begin{figure}[t]
    \centering
    \includegraphics[width=\linewidth]{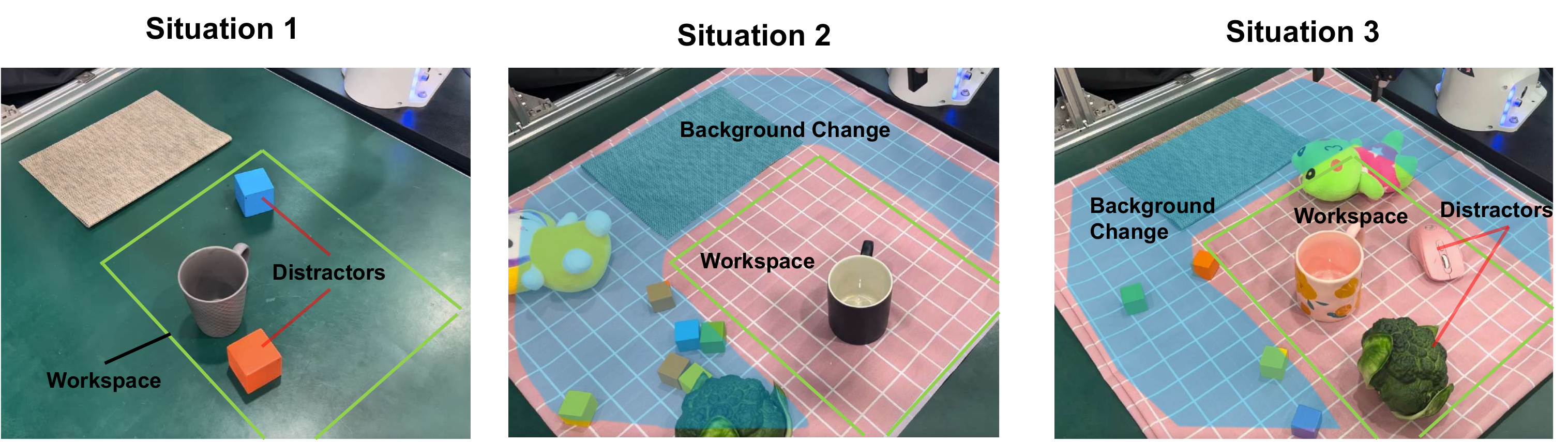}
    \caption{Visualization of three types of environment changes. \textit{Situation 1} adds distractors to the workspace, \textit{Situation 2} changes the background, and \textit{Situation 3} (the most difficult one) incorporates these two changes.}
    \label{fig:env-change}
\end{figure}

\paragraph{\textbf{Object generalization}} 
The results in Table \ref{tab:real-results} demonstrate that SKIL maintains remarkable performance even on unseen objects. In contrast, DP, DP3, and RISE perform poorly on unseen objects. 
GenDP-S performs slightly better than these three, by utilizing semantic fields to capture critical task-relevant information. However, SKIL uses semantic keypoint abstraction to obtain more concise and accurate representation. 
We take the \textit{Grasp Handle of Cup} task as an example.
As shown in Figure \ref{fig:keypoint-movement-objects} in Appendix \ref{app:keypoint-movement}, different cups exhibit varying appearances (shape, color, etc.), but share common structures (including the cup body and handle), which matter the most for this manipulation task. In practice, the semantic keypoint descriptors of SKIL effectively capture the information of these structures, while disregarding redundant details related to appearance, thus provides excellent generalization accross objects.


\begin{table}[t]
  \centering
  \caption{Average success rates of SKIL and baselines under the original \textit{Situation 0} and all three situations with environmental changes. SKIL outperforms the baselines by a large margin in all situations. }
  \begin{tabular}{@{}ccccc@{}}
    \toprule
     Method  & \textit{Situation 0} & \textit{Situation 1} & \textit{Situation 2} & \textit{Situation 3} \\
    \midrule
    DP & 4/10 & 4/10 & 0/10 & 0/10 \\
    DP3 & 5/10 & 1/10 & 4/10 & 0/10 \\
    RISE & 4/10 & 3/10 & 2/10 & 1/10 \\
    GenDP-S & 5/10 & 3/10 & 5/10 & 2/10 \\
    SKIL (Ours) & \textbf{9/10} & \textbf{9/10} & \textbf{8/10} & \textbf{8/10} \\
    \bottomrule
  \end{tabular}
  \label{tab:env-change}
\end{table}

\paragraph{\textbf{Environment generalization}}
We evaluate SKIL and other baselines on three new situations with environmental changes, including adding distractors (\textit{Situation 1}), background color shifting (\textit{Situation 2}) and their combination (\textit{Situation 3}), as illustrated in Figure \ref{fig:env-change}. For simplicity, we denote the original environment without any changes as \textit{Situation 0}.

We report the comparison results of the task \textit{Grasp Handle of Cup} in Table \ref{tab:env-change}, with the results of other tasks available in Appendix \ref{app:scene_general}. As shown, SKIL maintains consistently high performance across all situations. 
In contrast, baselines experience a substantial drop in performance especially in the most difficult \textit{Situation 3}.
Specifically, DP with image input suffers a severe performance drop when the background changes. DP3 is more resilient to background changes because it disregards color channels, but it performs poorly with additional distractors. GenDP performs better than other baselines but still suffers severe failures in \textit{Situation 3}. 
SKIL's semantic keypoints representation is least affected by environmental interference among these baselines, thus exhibiting superior generalization ability.
For a more detailed visualization, please refer to the supplementary video.

\begin{table*}[t]
  \centering
  \caption{Realworld results with different numbers of demonstrations. Here we use the two seen objects in the training phase to test the success rate, and conduct five trials for each object with random initialization in each task.}
  \begin{tabular}{@{}ccccccccccc@{}}
    \toprule
    \multicolumn{1}{l|}{Method/Task} & \multicolumn{2}{c|}{\textit{Pick Mouse}} & \multicolumn{2}{c|}{\textit{Grasp Handle of Cup}} & \multicolumn{2}{c|}{\textit{Fold Towel}} & \multicolumn{2}{c|}{\textit{Hang Towel}} & \multicolumn{2}{c}{\textit{Hang Cloth}} \\
    \multicolumn{1}{l|}{Number of Demos} & 10   & \multicolumn{1}{c|}{20} & 10 & \multicolumn{1}{c|}{20} & 10  & \multicolumn{1}{c|}{20} & 20 & \multicolumn{1}{c|}{30} & 20 & \multicolumn{1}{c}{30} \\
    \midrule
    DP & 20\% & 40\% & 10\% & 30\% & 40\% & 60\% & 0\% & 0\% & 0\% & 20\% \\
    DP3 & 10\% & 20\% & 20\% & 50\% & 50\% & 60\% & 0\% & 0\% & 0\% & 30\% \\
    RISE & 0\% & 10\% & 20\% & 40\% & 30\% & 50\% & 0\% & 0\% & 0\% & 20\% \\
    GenDP-S & 20\% & 40\% & 30\% & 50\% & 40\% & 60\% & 0\% & 0\% & 0\% & 30\% \\
    SKIL (Ours) & \textbf{50\%} & \textbf{90\%} & \textbf{70\%} & \textbf{90\%} & \textbf{60\%} & \textbf{90\%} & \textbf{40\%} & \textbf{70\%} & \textbf{40\%} & \textbf{60\%} \\
    \bottomrule
  \end{tabular}
  \label{tab:real-results-demos}
\end{table*}

\subsubsection{\textbf{Data Efficiency}}
Due to compounding errors, large amounts of data are indispensable for traditional imitation learning methods to get high performance, especially on long-horizon tasks. 
We consider the two long-horizon tasks, \textit{Hang Towel} and \textit{Hang Cloth}. These two tasks show different types of difficulty, one involves multiple pick-place actions, and the other requires precisely following a long spatial trajectory. 

Despite these challenges, SKIL reaches high success rates with only 30 demonstrations, outperforming all baselines by a large margin.
Particularly, SKIL achieves a success rate of 72\% on \textit{Hang Towel}, while all baselines fail completely. A prominent phenomenon is that they occasionally skip stages in the hanging process.
Appendix \ref{app:long-horizon} provides a detailed view of the task and illustrates the typical failure modes of the baselines.



Furthermore, we present the performance of SKIL and baselines with different numbers of demonstrations in Table \ref{tab:real-results-demos}. It can be seen that with the increase in demonstration amounts, SKIL's success rate grows much faster than the baselines. 
Specifically, the performance of SKIL with 10 demonstrations exceeds that of all baselines using 20 demonstrations on all tasks, showing SKIL's excelling data efficiency.


\begin{figure}[t]
    \centering
    \includegraphics[width=0.9\linewidth]{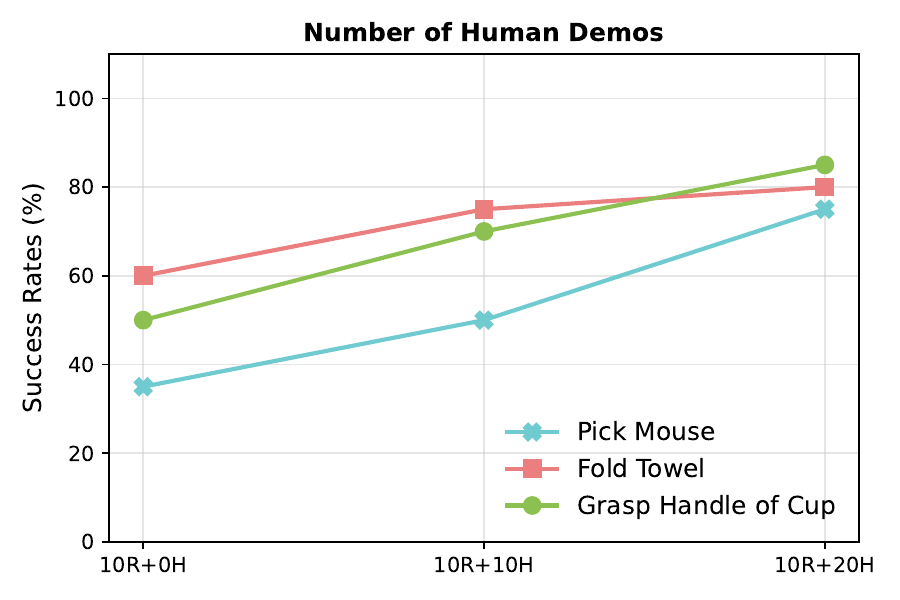}
    \caption{The performance comparison between different numbers of human videos without any action labels. ``10R+20H'' represents 10 robot demonstrations and 20 human demonstrations.}
    \label{fig:human_demos}
\end{figure}

\subsection{Cross-embodiment Performance}
By introducing a keypoint prediction model (Section \ref{sec:cross-embody}), SKIL-H enhances policy learning using extra human videos without action labels. We test SKIL-H on three tasks: \textit{Pick Mouse}, \textit{Grasp Handle of Cup}, and \textit{Fold Towel}, with 10 robot demonstrations and 0$\sim$20 human demonstrations. Figure \ref{fig:human-demo-vis} in Appendix \ref{app:human-demo} provides snapshots of human demonstration videos on these tasks. Quantitative results of final performance are shown in Figure \ref{fig:human_demos}. 
We can see that success rates increase significantly with the growth of human demo amounts. Particularly on the relatively hardest task \textit{Pick Mouse} among the three, 20 human demos lead to a dramatic 40\% increase in success rate, comparing to the policy trained solely on 10 robot demos. Besides, we also observe that with more human videos, SKIL-H produces smoother action sequences during evaluation. All these results show that the \textit{Trajectory Prediction Module} of SKIL-H do benifit from human videos, and further confirm the successful cross-embodiment semantic abstraction achieved by SKIL's keypoint description process. 



\begin{figure}[t]
    \centering
    \includegraphics[width=\linewidth]{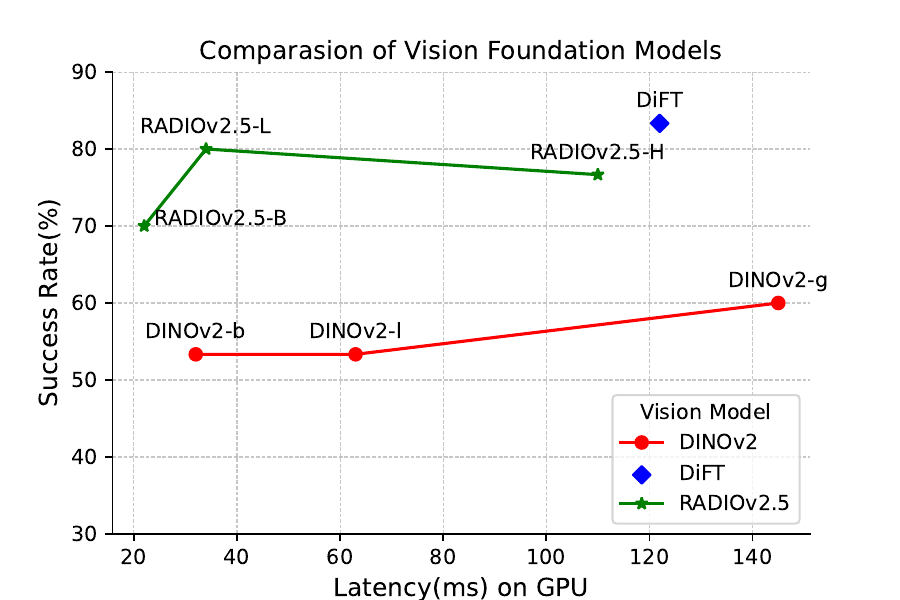}
    \caption{A comparison of success rates and inference latency for different vision foundation models (DINOv2, DiFT, and RADIO v2.5) on an NVIDIA A10 GPU. The results highlight the trade-off between computational overhead (latency) and performance (success rate) across varying model scales.}
    \label{fig:vlm-compare}
\end{figure}

\subsection{Ablations} \label{sec:ablation}
Since the core contribution of SKIL lies in the design of a novel representation for semantic keypoints, we conduct ablation studies to evaluate the impact of our design choices in selecting keypoints.
Specifically, we investigate the impact of different vision foundation models, keypoint numbers, and keypoint proposal strategies on three tasks: \textit{Pick Mouse}, \textit{Grasp Handle of Cup}, and \textit{Fold Towel}. 

\textbf{Ablation on Vision Foundation Models.}
In SKIL we use DiFT \cite{tang2023emergent} with Stable Diffusion 2.1 model, to extract features for later keypoint-related calculation, as described in Section \ref{sec:descriptor_kp}. We also tried 2 other recent models DINOv2 \cite{oquab2023dinov2} and RADIO \cite{ranzinger2024radio}, which are good at object detection and segmentation.
It can be seen from Figure \ref{fig:vlm-compare} that DINOv2 performs far behind the others, regardless of the size of backbone used. We observe that the keypoints obtained by DINOv2 suffer from severe mismatches, especially when objects are partially occluded, as illustrated in Figure \ref{fig:kp-vlm} in Appendix \ref{app:vision-models}.
On the other hand, the performance of RADIO models with ViT-L and ViT-H architectures are only slightly behind DiFT but with lower latencies, thus offering new choices for users to trade off performance and latency when implementing SKIL in specific scenes. 
Note that SKIL itself does not rely on any specific vision foundation model, so we believe that it could continue to benefit from any latest model to get even higher performance in the future. More details can be found in Appendix \ref{app:vision-models}.

\begin{table}[t]
  \centering
  \caption{Average success rates with different keypoint numbers ($N$).}
  \begin{tabular}{@{}cccc@{}}
    \toprule
     Num. keypoints ($N$) & \textit{Pick Mouse} & \textit{Fold Towel} & \textit{Grasp Handle of Cup} \\
    \midrule
    10 & 7/10 & 7/10 & 7/10 \\
    20  & \textbf{8/10} & \textbf{8/10} & 8/10 \\
    30  & \textbf{8/10} & 7/10 & \textbf{9/10} \\
    \bottomrule
  \end{tabular}
  \label{tab:num_keypoints}
\end{table}

\begin{table}[t]
  \centering
  \caption{Average success rates with different keypoint proposal strategies.   ``Random'' means selecting keypoints randomly inside the object mask, and ``Manual'' means manually selecting keypoints based on human knowledge. We choose to use K-means for SKIL. }
  \begin{tabular}{@{}cccc@{}}
    \toprule
     Method  & \textit{Pick Mouse} & \textit{Fold Towel} & \textit{Grasp Handle of Cup} \\
    \midrule
    Manual & 7/10 & \textbf{7/10} & \textbf{9/10} \\
    Random  & 7/10 & 6/10 & \textbf{9/10}\\
    K-means(Ours)  & \textbf{8/10} & \textbf{7/10} & \textbf{9/10} \\
    \bottomrule
  \end{tabular}
  \label{tab:kp-select-methods}
\end{table}


\textbf{Ablation on Keypoint Numbers.}
We investigate the impact of the number of keypoints ($N$) set in the \textit{Semantic Keypoints Description Module} (Section \ref{sec:descriptor_kp}). Experiment results are shown in Table \ref{tab:num_keypoints}, which illustrate that SKIL achieves similar performance with 10-30 keypoints. That means the performance is insensitive to the change of keypoint numbers in this range.
Actually, higher $N$ leads to higher dimensionality of keypoint descriptors, which means more information encoded with higher computing cost.
Future users of SKIL may easily find appropriate values of $N$ with few trials to reach enough performance with the least computing cost on new tasks.

\textbf{Ablation on Keypoint Proposal Strategies.}
Keypoint proposal aims to identify the reference keypoints on the target objects from the reference image. We choose to use K-means for SKIL as illustrated in Section \ref{sec:descriptor_kp}.
Here, we compare its effectiveness with 2 other strategies (selecting keypoints manually or randomly on objects in the reference image). As shown in Table \ref{tab:kp-select-methods}, these 2 strategies achieve very similar performance, with the random strategy slightly inferior only on the \textit{Fold Towel} task. We speculate that the lack of keypoints on edge of the towel hinders accurate grasping of the edge. Our choice K-means performs best among these strategies on all tasks, with its strong clustering ability of object features and independence of human inductive bias.

\section{Limitations}


Although SKIL has demonstrated extraodinary performance in these manipulation tasks, its capability is strictly upper-bounded by the capability of its upstream vision foundation model. As an example, we have tried but struggled to complete a \textit{Bulb Assembly} task with SKIL, because the precision of keypoints extracted by the current model (DiFT) could not reach the high requirement of such task. Another limitation is that current SKIL is unable to complete tasks that require detailed perception of environments, as it only extracts keypoints from the relevant objects. For instance, it might violate safety constraints on tasks with multiple obstacles. Future work may extend the capability of SKIL by developing an efficient keypoint-based environment representation.


\section{Conclusions}
High sample complexity remains a significant barrier to advancing imitation learning for generalizable and long-horizon tasks. To address this challenge, we develop the Semantic Keypoints Imitation Learning (SKIL) algorithm.
Leveraging a vision foundation model, SKIL obtains the semantic keypoints as sparse observations, significantly reducing the dimensionality of the problem, and the proposed descriptors of semantic keypoints substantially improve the policy's generalization ability. Furthermore, the semantic keypoint abstraction of SKIL naturally supports cross-embodiment learning. 
Experiments demonstrate that SKIL achieves excelling data efficiency and strong generalization ability.
We believe that our work can pave the way for the development of general-purpose robots capable of solving complicated open-world problems.




\bibliographystyle{plainnat}
\bibliography{references}

\clearpage
\appendix

\subsection{Simulation Environment}
\label{app:simenv}

\textbf{Benchmarks.} 
Metaworld consists of 50 distinct robotic manipulation tasks using a sawyer robot. The four-dimensional action space includes the relative changes in the end-effector position and gripper state, while observations consist of an RGB image and the corresponding depth image. Relative changes in the end-effector position range from –1 to 1, and the gripper state ranges from 0 to 1.

The DexArt benchmark consists of 4 dexterous manipulation tasks. The action space is 22-dimensional because DexArt employs a 16-DoF Allegro hand and a 6-DoF Xarm. Each dimension of the action space represents the relative change in joint position, ranging from -1 to 1. Besides, the observations are the same as those in MetaWorld. Notably, DexArt uses different objects during the training and evaluation phases. 

\begin{figure}[ht]
  \centering
  \includegraphics[width=\linewidth]{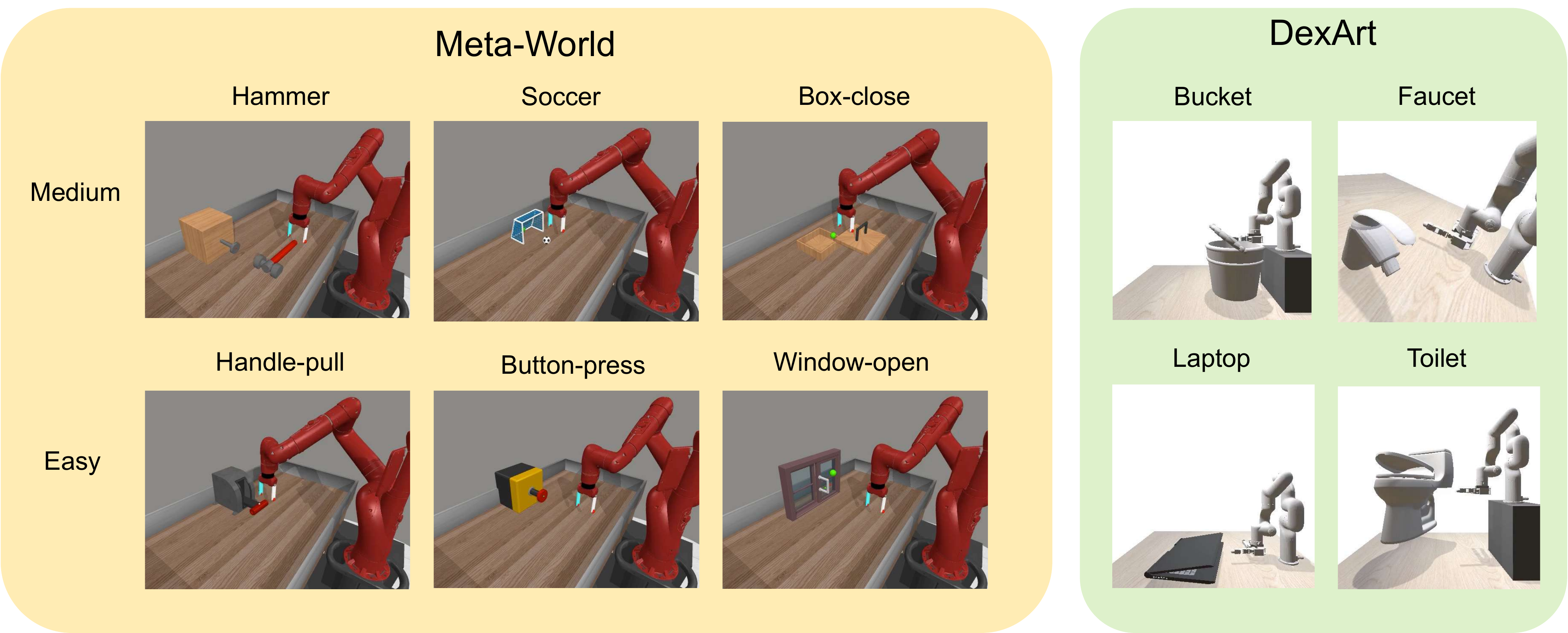}
  \caption{Specific simulation tasks used in the MetaWorld and DexArt benchmarks.}
  \label{fig:simulation}
\end{figure}

\textbf{Tasks.} For the simulation experiments, we select 6 tasks from the MetaWorld \cite{yu2020meta} and all 4 tasks in DexArt \cite{bao2023dexart}, as shown in Figure \ref{fig:simulation}. The tasks in MetaWorld are categorized into different difficulty levels based on the criteria in \cite{seo2023masked}. We chose three easy-level tasks and three medium-level tasks. 

\textbf{Training Details.}
We collect expert demonstrations using scripted policies in MetaWorld, and reinforcement learning (RL) agents in DexArt. For each task, we collect 10 demonstrations and train the policies using 3 random seeds. During training, we evaluate the policies every 100 epochs over 10 episodes and report the average of the highest 3 success rates. The final performance is reported as the mean and standard deviation of the success rates across the 3 seeds.

\subsection{Real-world Environment Setup}

\subsubsection{Hardware}
\label{app:hardware}
\begin{figure}[ht]
  \centering
  \includegraphics[width=0.9\linewidth]{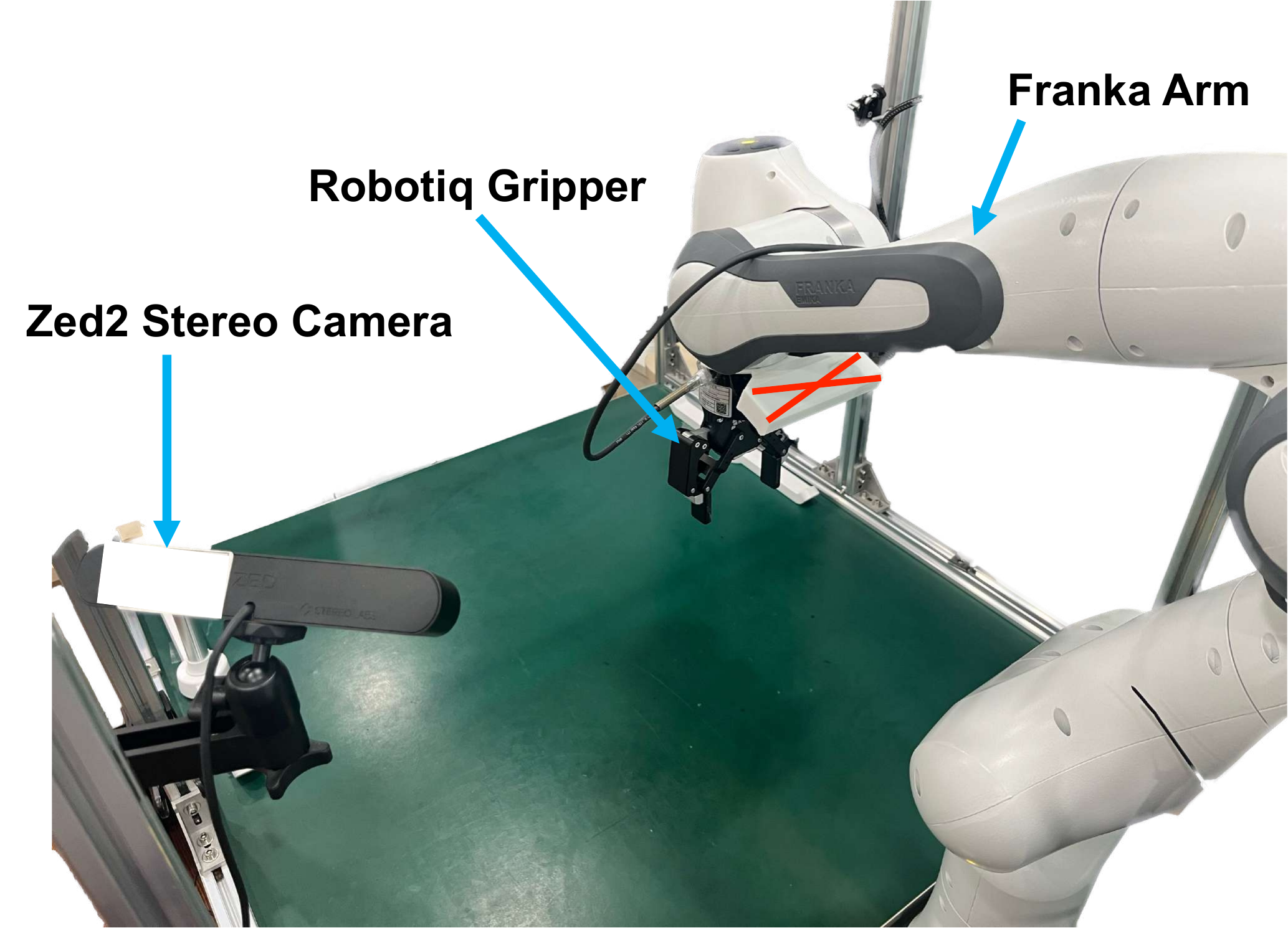}
  \caption{Real-world hardware setup. We use a Franka arm equipped with a Robotiq gripper. A fixed Zed2 stereo camera is employed to capture visual and depth observations. Note that the wrist camera is not used in our experiments.}
  \label{fig:hardware}
\end{figure}

Our real-world setup is shown in Figure \ref{fig:hardware}. We primarily follow the hardware configuration of the Droid dataset \cite{khazatsky2024droid}. As depicted in Figure \ref{fig:hardware}, a wrist camera is mounted on the Franka arm similar to the Droid setup; however, we do not use the wrist camera throughout our experiments.

\subsubsection{Random Initialization}
\label{app:randomization}
\begin{figure}[t]
  \centering
  \includegraphics[width=0.9\linewidth]{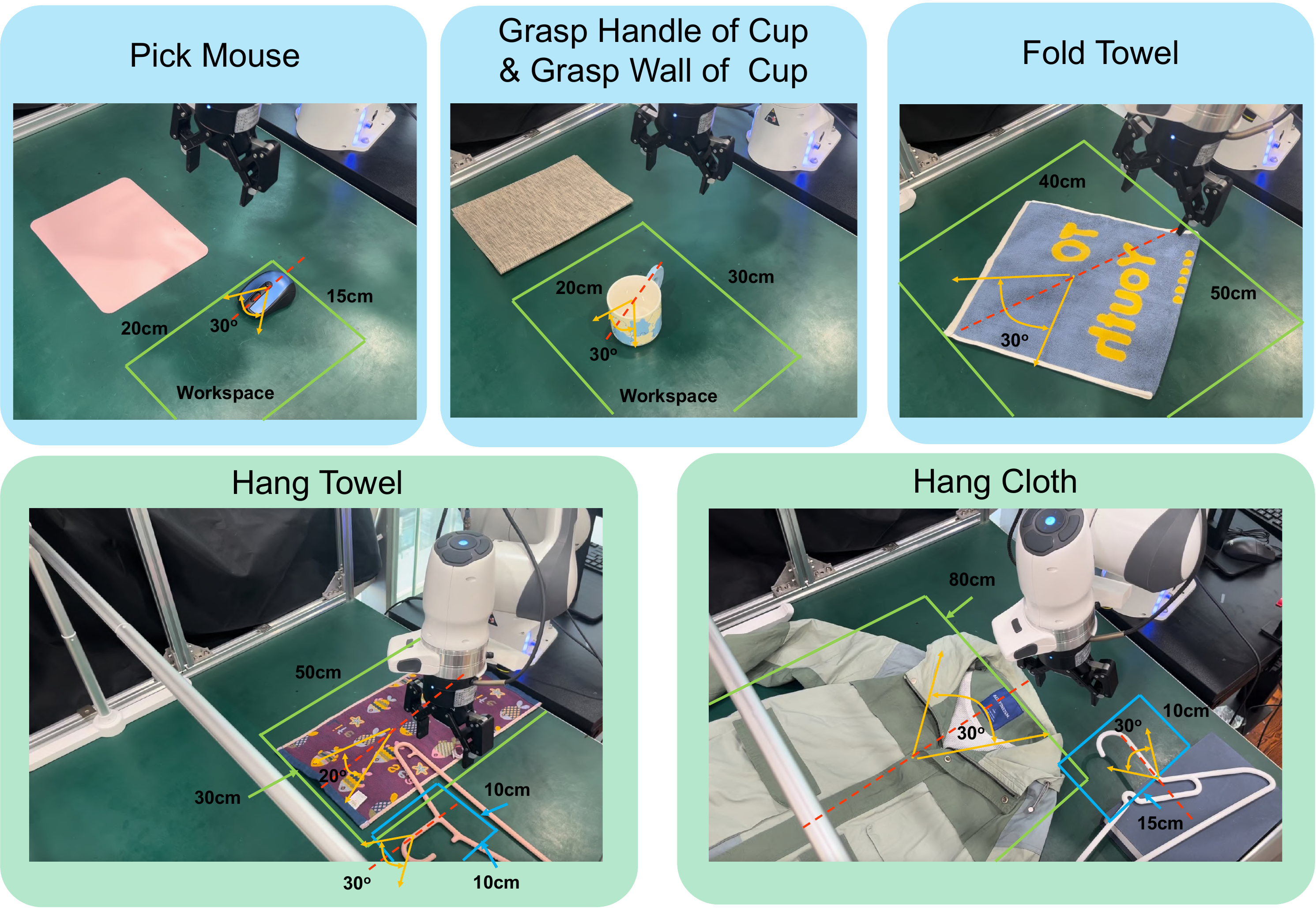}
  \caption{Workspace of the six real-world tasks, in which we account for random positions and orientations of objects. The green and blue squares indicate the workspace, and the orange fan-shaped area represents the range of random orientations.}
  \label{fig:randomization}
\end{figure}

The workspaces for all real-world tasks are shown in Figure \ref{fig:randomization}. We consider the random positions and orientations of objects in each task. Note that the blue square represents the working space of the hanger hook in the two long-horizon tasks, \textit{Hang Towel} and \textit{Hang Cloth}. Throughout data collection and evaluation, 
we put each object at a random position inside the workspace, with a random orientation in a certain range.

\subsection{Visualization of Real-world Task Objects}
\label{app:objects}
\begin{figure*}[ht]
  \centering
  \includegraphics[width=\linewidth]{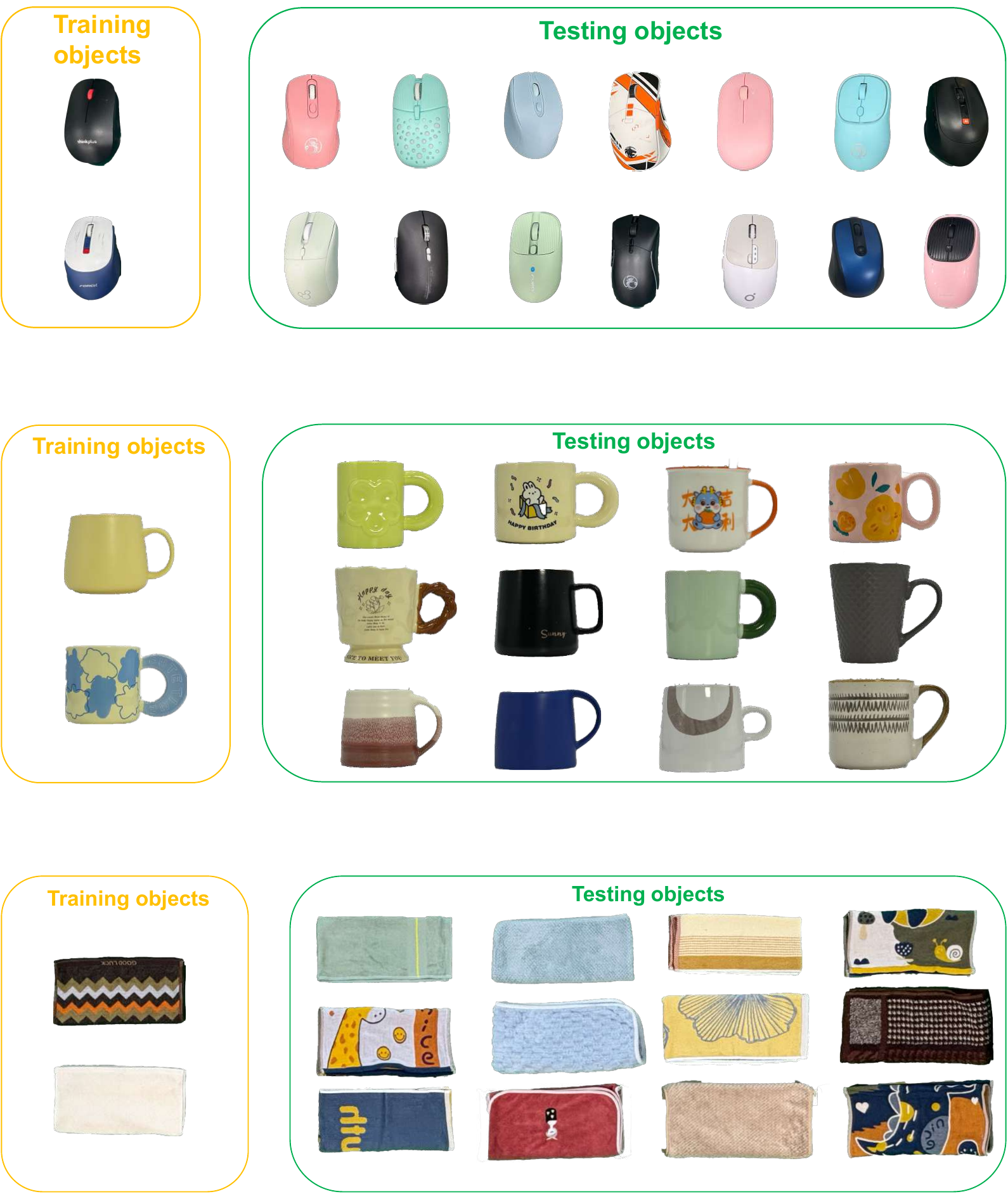}
  \caption{Visualization of training and testing objects. We use 2 objects for training and 10 objects for testing on short-horizon tasks, including \textit{Pick Mouse}, \textit{Grasp Handle of Cup}, \textit{Grasp Wall of Cup}, and \textit{Fold Towel}.}
  \label{fig:objects-1}
\end{figure*}

\begin{figure*}[ht]
  \centering
  \includegraphics[width=\linewidth]{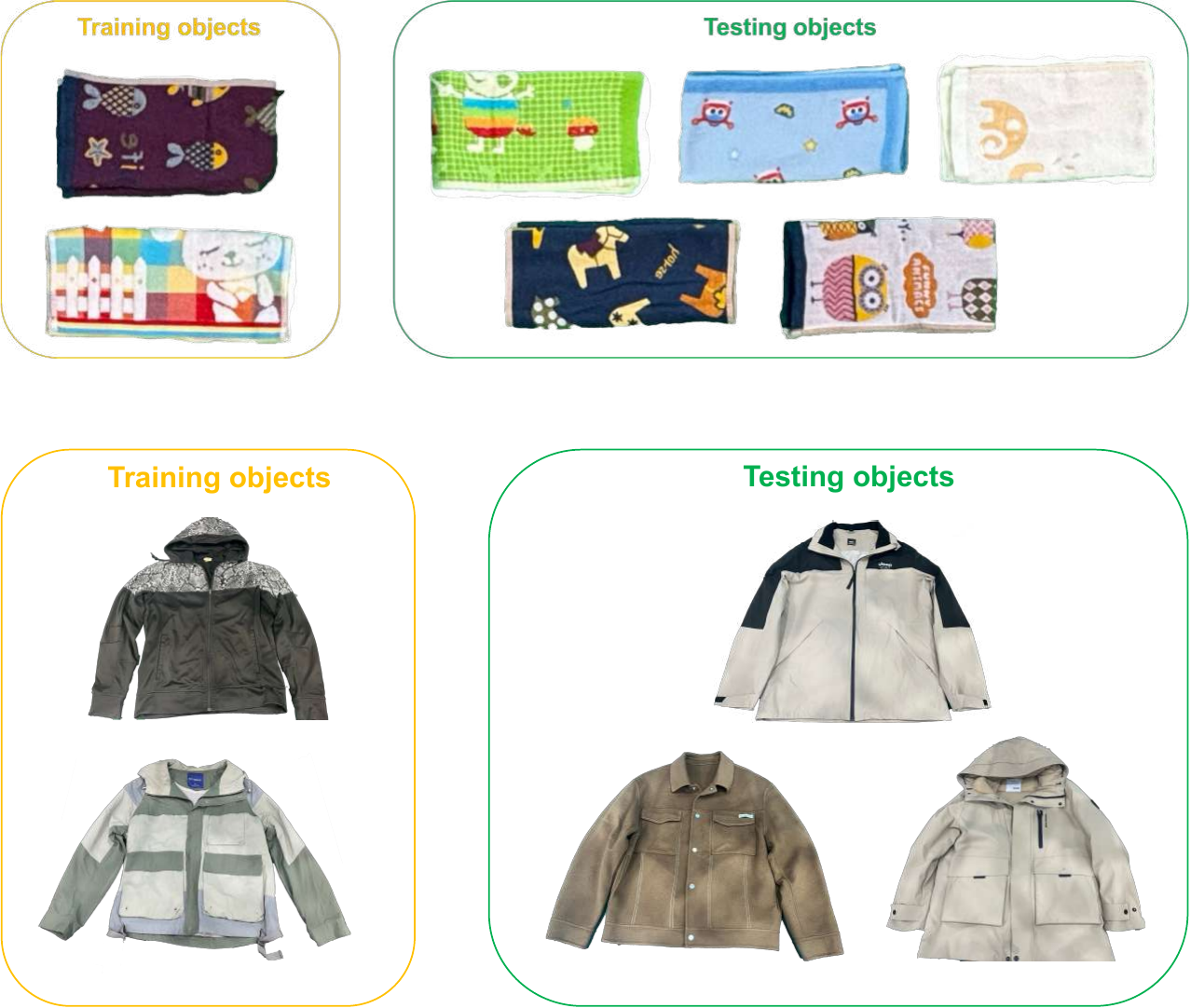}
  \caption{Visualization of training and testing objects. We use 2 objects for training and 3 or 5 objects for testing on the long-horizon tasks, including \textit{Hang Towel} and \textit{Hang Cloth}.}
  \label{fig:objects-2}
\end{figure*}

In Figures \ref{fig:objects-1} and \ref{fig:objects-2}, we show the training and testing objects for all the real-world tasks \textit{Pick Mouse}, \textit{Grasp Handle of Cup}, \textit{Grasp Wall of Cup}, \textit{Fold Towel}, \textit{Hang Towel}, and \textit{Hang Cloth}, respectively.

\subsection{Details of Baselines}
\label{app:baselines}
We summarize the baselines as follows:
\begin{enumerate}
    \item Diffusion Policy (DP) \cite{chi2023diffusion}: This method models the action distribution using a diffusion model, leveraging RGB observations as conditions within the diffusion process to generate robot actions.
    \item DP3 \cite{ze20243d}: DP3 adopts a similar diffusion architecture to DP but introduces a compact 3D representation instead of using 2D images, leveraging an efficient MLP encoder for 3D data processing.
    \item RISE \cite{wang2024rise}: RISE uses 3D point clouds to predict robot actions by processing the point cloud data with a shallow 3D encoder, which is then mapped to actions using a transformer-based network.
    \item GenDP-S \cite{wang2024gendp}: GenDP generates 3D descriptor fields from multi-view RGBD data and computes semantic fields by measuring the cosine similarity between these descriptors and 2D reference features. These semantic fields are then combined with the point cloud data and passed through PointNet++ and a diffusion policy to predict robot actions. Note that since we construct 3D descriptor fields using only a single camera, we refer to our implementation as GenDP-S.
\end{enumerate}
Since all baselines use a diffusion-based action head, we maintain the same diffusion parameters as in our method, SKIL. For other parameters, we adhere to the settings outlined in the respective papers.

\subsection{Simulation Experiment Results}
\label{app:sim-results}

Results of SKIL and baselines on simulation tasks are shown in Table \ref{tab:metaworld-results} and \ref{tab:dexart-results}. We see that SKIL
outperforms all baselines on all simulation tasks, but with smaller leading gaps than on real-world tasks. We list here the possible reasons we believe:
\begin{itemize}
    \item Simulators always provide perfect observations, which lead to easier state abstraction and thus reduces the advantage of SKIL in this aspect.
    \item Most MetaWorld tasks are much simpler than real-world tasks, so that even the baselines could get high performance.
    \item The action dimension of DexArt tasks (22-Dof) are much higher than that of other tasks, so the much larger action space limits the performance of both SKIL and the baselines on these tasks..
\end{itemize}


\begin{table}[ht]
  \centering
  \caption{DexArt Results. We present the mean and standard deviation (std) of the success rates for each task.}
  \begin{tabular}{@{}ccccc@{}}
    \toprule
    \multicolumn{1}{l|}{Method/Task} & \multicolumn{1}{c|}{\textit{Bucket}} & \multicolumn{1}{c|}{\textit{Toilet}} & \multicolumn{1}{c|}{\textit{Faucet}} & \multicolumn{1}{c}{\textit{Laptop}} \\
    \midrule
    DP & $54.4${\scriptsize(15.0)} & $36.7${\scriptsize(12.5)} & $25.5${\scriptsize(6.8)} & $28.9${\scriptsize(1.6)} \\
    DP3 & $28.9${\scriptsize(5.7)} & \textbf{45.5}{\scriptsize(5.6)} & $14.4${\scriptsize(3.1)} & $38.9${\scriptsize(6.9)} \\
    RISE & $52.2${\scriptsize(12.2)} & $34.4${\scriptsize(10.3)} & $11.1${\scriptsize(1.6)} & $27.8${\scriptsize(12.8)} \\
    GenDP-S & $28.9${\scriptsize(1.6)} & $32.2${\scriptsize(4.2)} & $24.4${\scriptsize(6.3)} & $43.3${\scriptsize(2.7)} \\
    SKIL(Ours) & \textbf{61.6}{\scriptsize(1.6)} & \textbf{45.5}{\scriptsize(5.6)} & \textbf{27.7}{\scriptsize(4.1)} & \textbf{44.4}{\scriptsize(4.2)} \\
    \bottomrule
  \end{tabular}
  \label{tab:dexart-results}
\end{table}

\begin{table*}[ht]
  \centering
  \caption{Metaworld Results. We present the mean and standard deviation (std) of the success rates for each task.}
  \begin{tabular}{@{}ccccccc@{}}
    \toprule
    \multicolumn{1}{l|}{Method/Task} & \multicolumn{1}{c|}{\textit{Hammer}} & \multicolumn{1}{c|}{\textit{Handle-pull}} & \multicolumn{1}{c|}{\textit{Soccer}} & \multicolumn{1}{c|}{\textit{Box-close}} & \multicolumn{1}{c|}{\textit{Button-press}} & \multicolumn{1}{c|}{\textit{Window-open}}\\
    \midrule
    DP & $54.3${\scriptsize(6.1)} & $18.9${\scriptsize(4.2)} & $21.1${\scriptsize(4.2)} & $53.3${\scriptsize(2.7)} & \textbf{100}{\scriptsize(0)} & $85.5${\scriptsize(3.2)} \\
    DP3 & $71.1${\scriptsize(4.1)} & $62.2${\scriptsize(3.2)} & $5.3${\scriptsize(3.3)} & $70.3${\scriptsize(2.3)} & \textbf{100}{\scriptsize(0)} & \textbf{100}{\scriptsize(0)} \\
    RISE & $36.7${\scriptsize(12.5)} & $17.8${\scriptsize(3.1)} & $11.1${\scriptsize(1.6)} & $43.3${\scriptsize(9.8)} & \textbf{100}{\scriptsize(0)} & $58.9${\scriptsize(6.3)} \\
    GenDP-S & $57.8${\scriptsize(5.7)} & $31.1${\scriptsize(6.9)} & $7.7${\scriptsize(1.6)} & $51.1${\scriptsize(3.1)} & $72.2${\scriptsize(5.7)} & $46.7${\scriptsize(5.4)} \\
    SKIL(Ours) & \textbf{100}{\scriptsize(0)} & \textbf{75.6}{\scriptsize(4.2)} & \textbf{24.4}{\scriptsize(3.1)} & \textbf{71.1}{\scriptsize(8.7)} & \textbf{100}{\scriptsize(0)} & \textbf{100}{\scriptsize(0)} \\
    \bottomrule
  \end{tabular}
  \label{tab:metaworld-results}
\end{table*}

\subsection{Real-world Experiment Results}

\subsubsection{Visulization of Semantic Keypoints}
\label{app:keypoint-movement}
\begin{figure*}[ht]
  \centering
  \includegraphics[width=\linewidth]{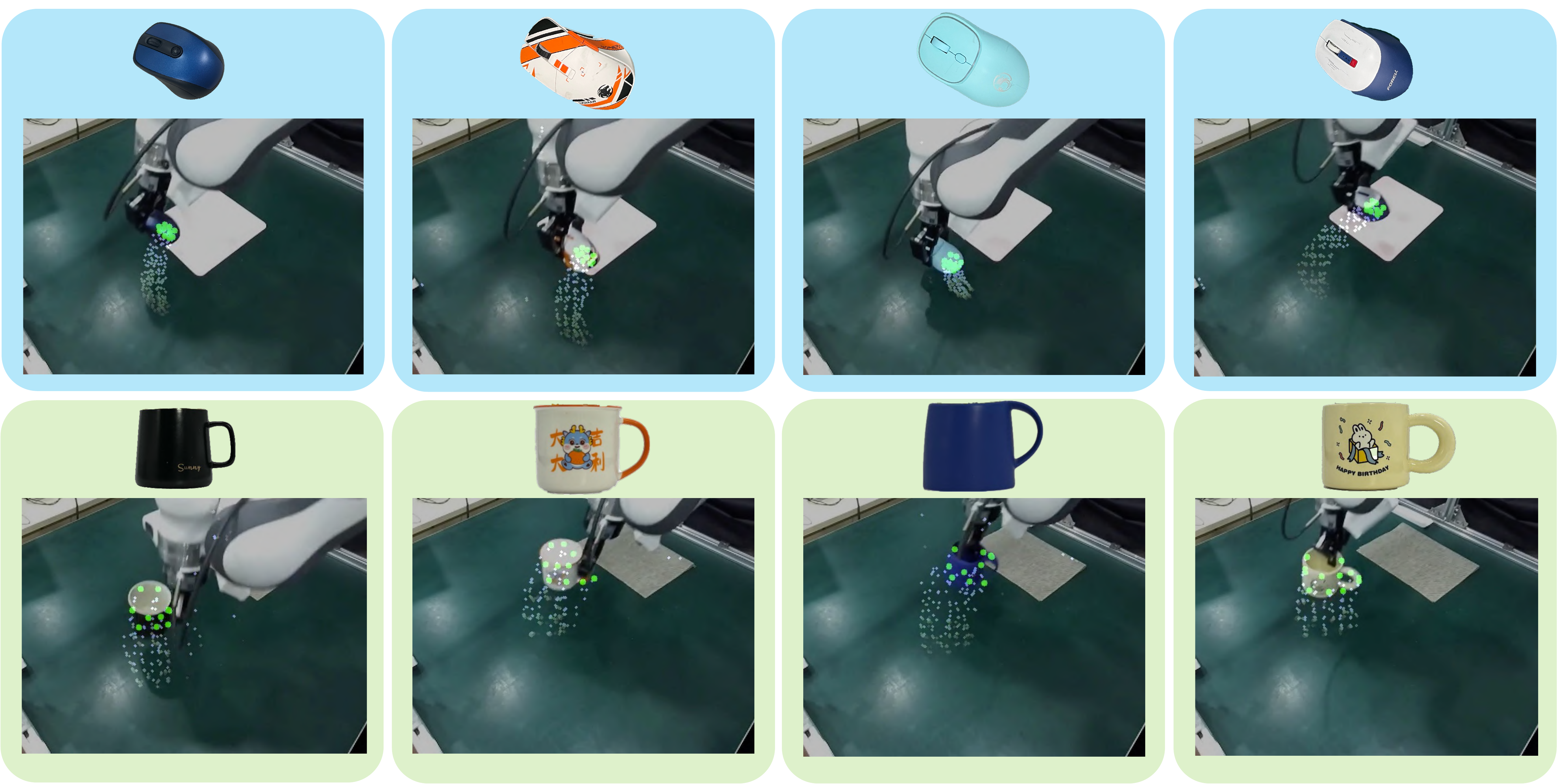}
  \caption{Movement of semantic keypoints on different objects in SKIL. Green points represent the current keypoints, and the white flows show their previous trajectories. Temporal consistency is maintained across objects within the same category.}
  \label{fig:keypoint-movement-objects}
\end{figure*}

We visualize the semantic keypoints on different objects, as shown in Figure \ref{fig:keypoint-movement-objects}. The results demonstrate that temporal consistency is preserved across objects within the same category.




\subsubsection{Analysis of Grasp Wall of Cup}
\label{app:cupwall_analysis}

Table \ref{tab:real-results} shows the baselines with point clouds input (e.g. DP3, RISE and GenDP-S) perform poorly on \textit{Grasp Wall of Cup}. In contrast, DP achieves a success rate of 70\% on the training objects.
We observe that the task requires grasping the inner wall of cup on the side away from the camera, but DP3, RISE and GenDP-S always perform grasping at a small distance from the cup. We suspect that the Zed2 stereo camera produces low-quality point clouds around the inner wall of the cups, due to the pure white color of that part.
Thus, these methods with point clouds input observe the inaccurate positions of the cup wall. 
However, SKIL fuses the information of all keypoints using a transformer encoder, so that the perception errors on these small fraction of keypoints can be eliminated by their neighbours to some extent.

\subsubsection{Spatial Generalization of SKIL} 
\label{app:spatial_gen}
\begin{figure}[ht]
  \centering
  \includegraphics[width=\linewidth]{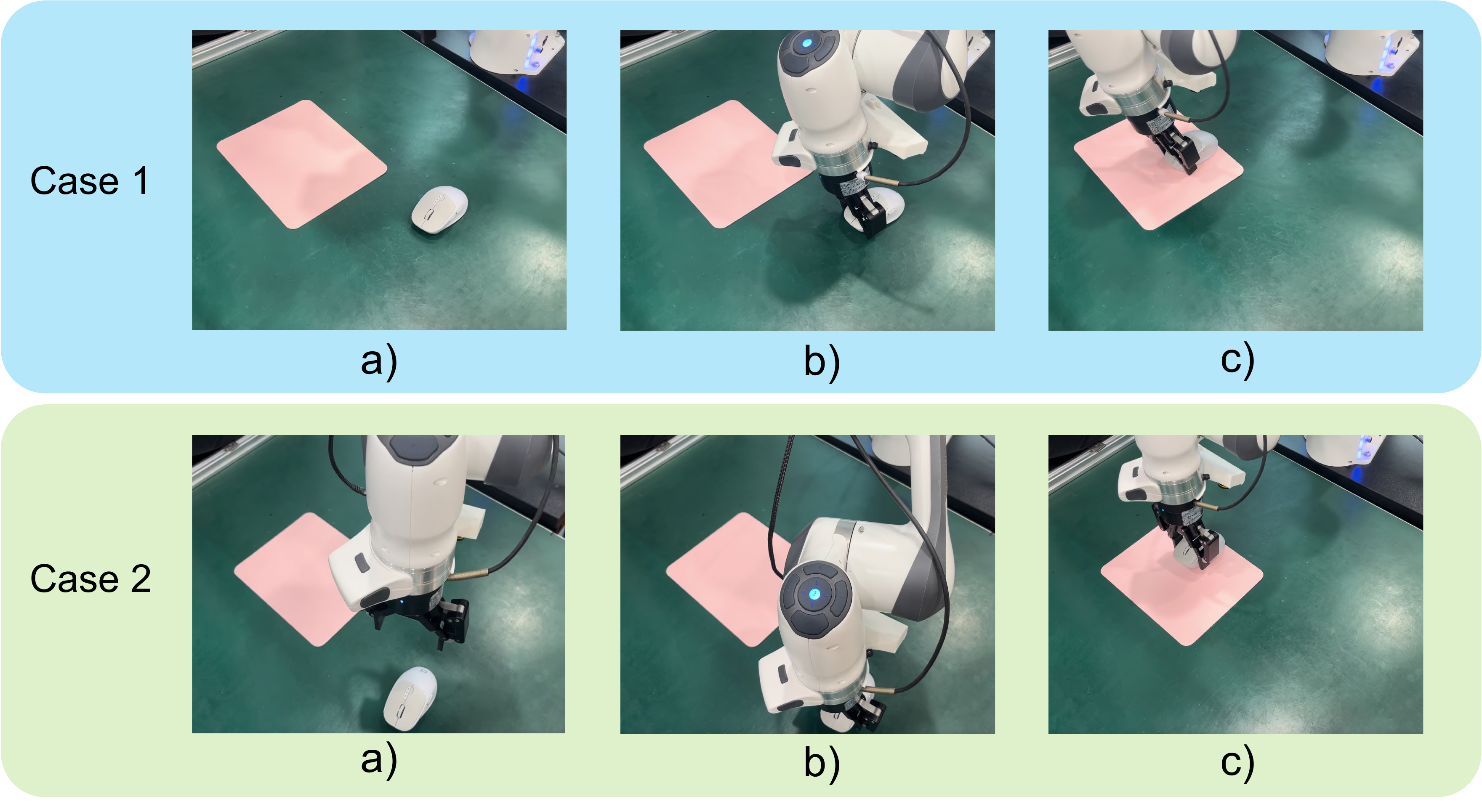}
  \caption{SKIL successfully handles objects positioned at the edge of the workspace. We demonstrate its ability to grasp a mouse located in the corner of the workspace.}
  \label{fig:mouse-corner}
\end{figure}

\begin{figure}[ht]
  \centering
  \includegraphics[width=\linewidth]{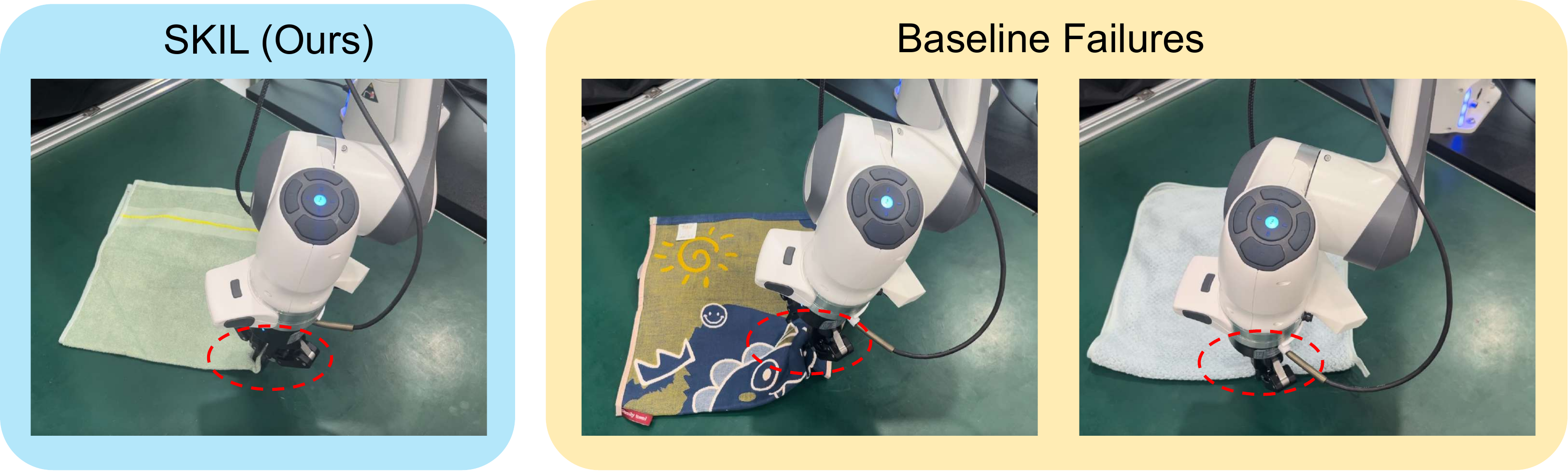}
  \caption{SKIL accurately grasps the corner of the towel, whereas baselines struggle with precise grasping.}
  \label{fig:towel-failure}
\end{figure}

We find that our method can effectively handle objects located at the edges of the workspace, as shown in Figure \ref{fig:mouse-corner}. In contrast, baselines often fail to handle objects positioned at the corners of the workspace, especially when dealing with unseen objects.
As shown in Figure \ref{fig:towel-failure}, the SKIL method can precisely grasp the left corner of the towel, as demonstrated during the training. In contrast, while baselines can fold the towel, their policies often grasp the wrong region, missing the left corner.

\subsubsection{Environment generalization of SKIL}
\label{app:scene_general}
Other than \textit{Grasp Handle of Cup}, we also conduct experiments on environment generalization for \textit{Pick Mouse} and \textit{Fold Towel}. As shown in Table \ref{tab:pick-mouse-scene} and \ref{tab:fold-towel-scene}, all algorithms exhibit similar performance across these tasks. Therefore, we omit further analysis for \textit{Pick Mouse} and \textit{Fold Towel}. For \textit{Hang Towel}, a long-horizon task, while all baselines fail to complete the task, SKIL achieves high performance under all situations.

\subsubsection{Long-horizon manipulation of SKIL} 
\label{app:long-horizon}
\begin{figure}[ht]
  \centering
  \includegraphics[width=\linewidth]{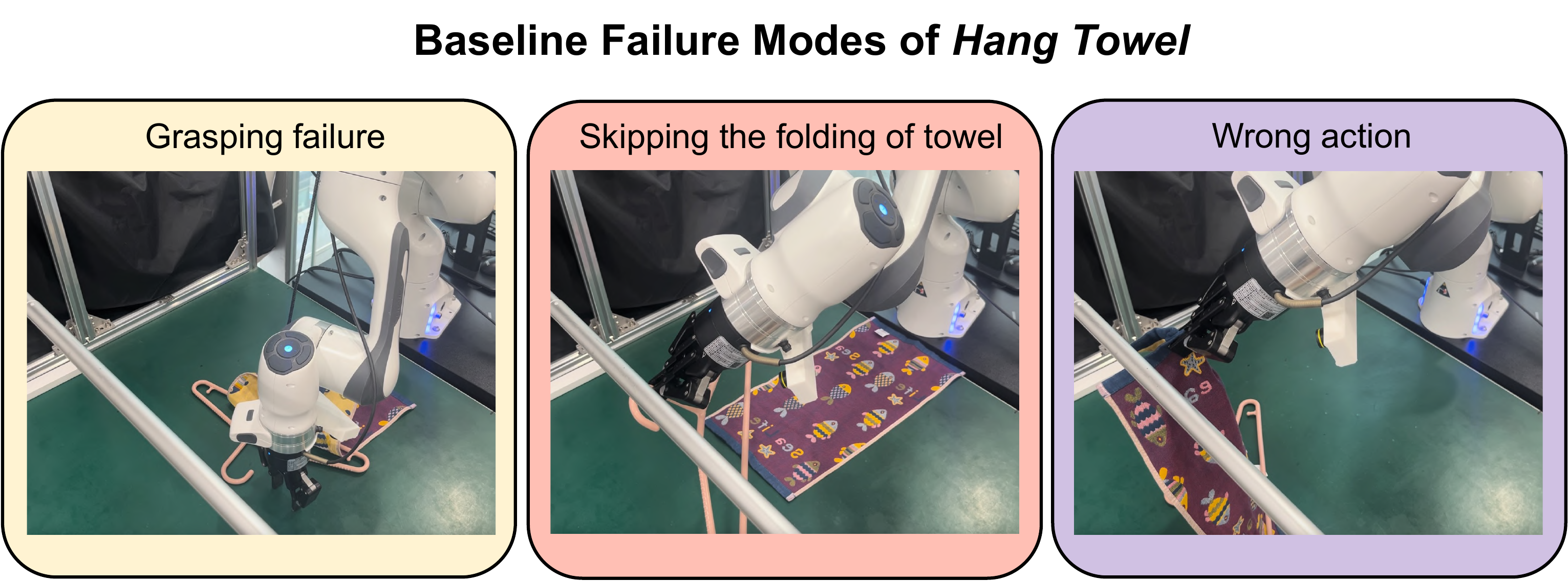}
  \caption{Classical failure modes of baselines in \textit{Hang Towel}, including grasping failure, skipping the folding of towel, and wrong action (directly putting the towel onto the rack without the hanger).}
  \label{fig:hangtowel-failure}
\end{figure}

\begin{figure*}[ht]
  \centering
  \includegraphics[width=\linewidth]{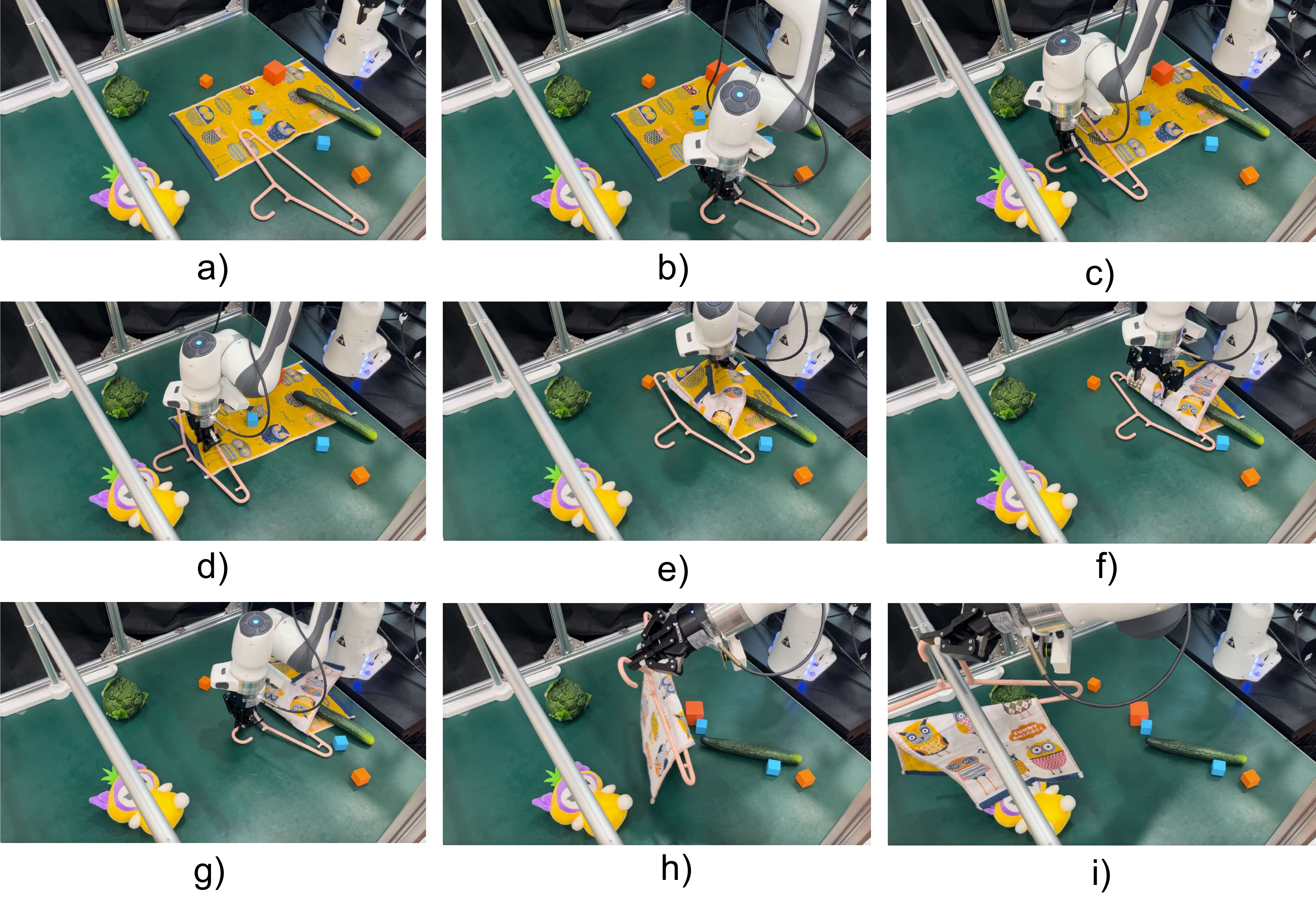}
  \caption{Videosnaps of \textit{Hang Towel} using SKIL. We can see that this task includes multiple stages: grasping the handle of hanger (a-b), placing the hanger on the towel's edge (b-c), grasping and folding the towel (d-f), grasping the handle hook (f-g), and finally hanging it on the rack (h-i).}
  \label{fig:long-horizon-towel}
\end{figure*}

\begin{figure*}[ht]
  \centering
  \includegraphics[width=\linewidth]{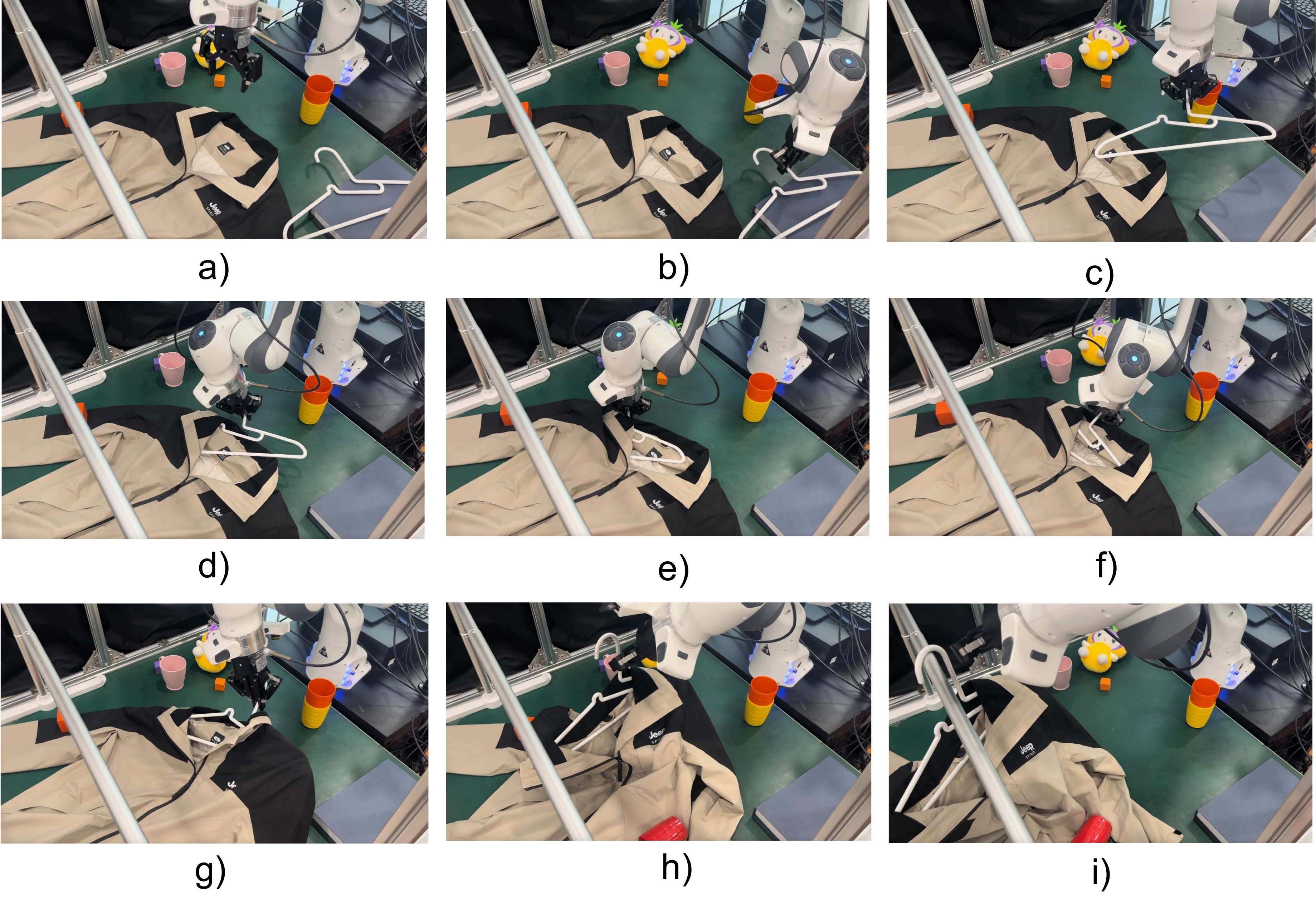}
  \caption{Videosnaps of \textit{Hang Cloth} using SKIL. Notice that the cloth and scene are unseen for the policy. The detailed stages include grasping the handle of hanger (a-b), inserting the hanger into the cloth (c-f), and hanging it onto the rack (g-i).}
  \label{fig:long-horizon-cloth}
\end{figure*}

Figure \ref{fig:long-horizon-towel} and \ref{fig:long-horizon-cloth} illustrate the whole process of \textit{Hang Towel} and \textit{Hang Cloth} on a rack. Notice that baselines perform poorly on \textit{Hang Towel}, and we show the classical failure mode of baselines in Figure \ref{fig:hangtowel-failure}. For example, the gripper cannot grasp the handle hook of hanger after folding the towel. This is because the position of the handle hook is disturbed behind folding the towel, as shown in Figure \ref{fig:long-horizon-towel} (d, e, f). Due to limited demonstrations, baselines are unable to generalize to different positions of the handle. Besides, baselines occasionally skip necessary actions like folding the towel because of perception errors. 
Sometimes, wrong actions also occur. For example, the robot may put the towel directly onto the rack without the hanger.
To conclude these, most failures result from high compounding errors in long-horizon tasks. 


\subsubsection{Visualization of human videos}
\label{app:human-demo}
\begin{figure*}[ht]
  \centering
  \includegraphics[width=\linewidth]{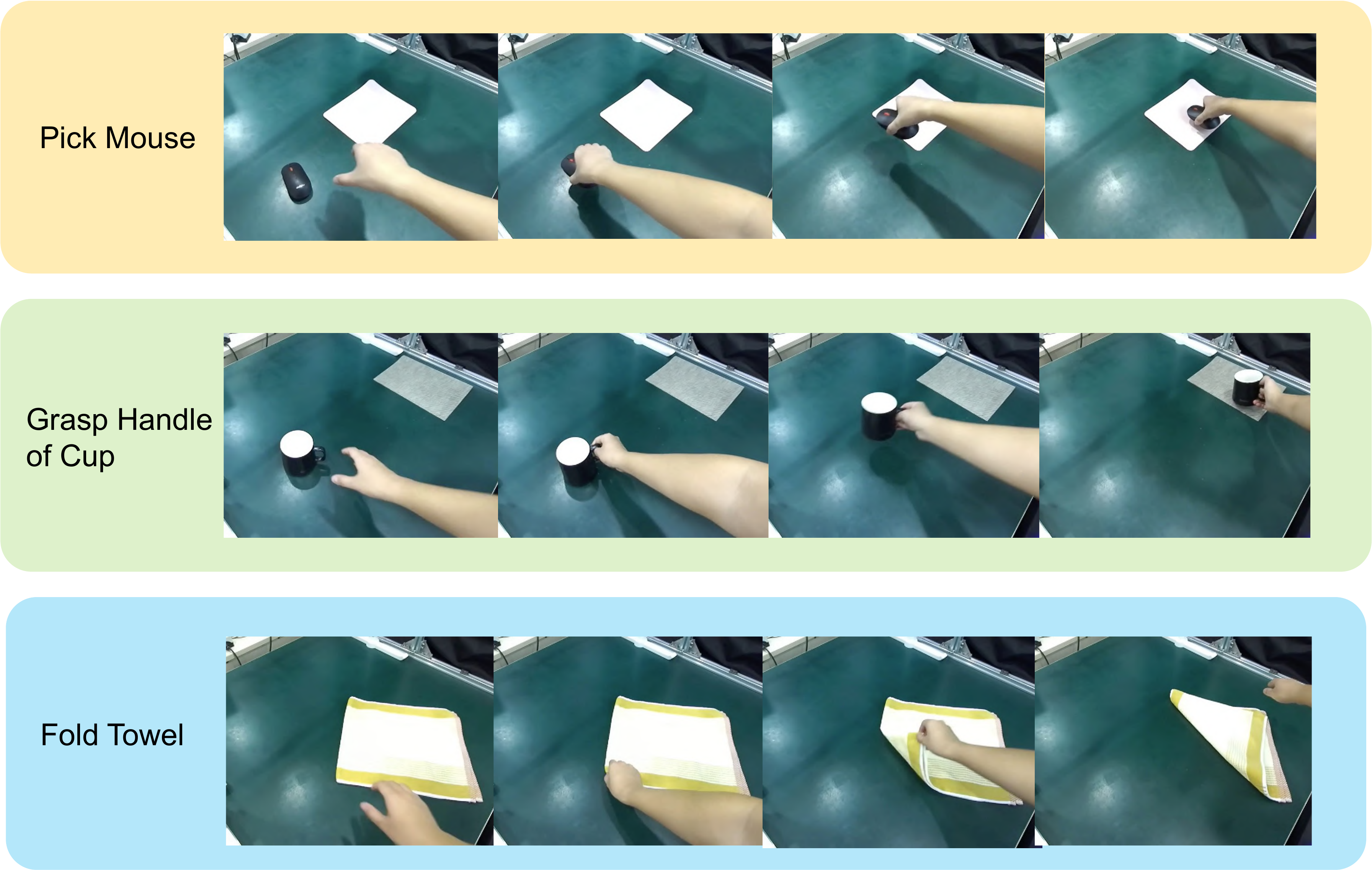}
  \caption{Videosnaps of human demos collected in \textit{Pick Mouse}, \textit{Grasp Handle of Cup}, and \textit{Fold Towel}.}
  \label{fig:human-demo-vis}
\end{figure*}

We visualize the human videos in \textit{Pick Mouse}, \textit{Grasp the Handle of Cup}, and \textit{Fold Towel}, as shown in Figure \ref{fig:human-demo-vis}.

\begin{table}[ht]
  \centering
  \caption{Average success rates of SKIL and baselines on the \textit{Pick Mouse} task under three types of environmental changes. (\textit{Situation 0} means without any environmental changes.)}
  \begin{tabular}{@{}ccccc@{}}
    \toprule
    Method & \textit{Situation 0} & \textit{Situation 1 } & \textit{Situation 2} & \textit{Situation 3} \\
    \midrule
    DP & 2/10 & 2/10 & 0/10 & 0/10 \\
    DP3 & 3/10 & 1/10 & 3/10 & 0/10 \\
    RISE & 1/10 & 0/10 & 1/10 & 0/10 \\
    GenDP-S & 4/10 & 3/10 & 0/10 & 0/10 \\
    SKIL (Ours) & \textbf{7/10} & \textbf{7/10} & \textbf{6/10} & \textbf{5/10} \\
    \bottomrule
  \end{tabular}
  \label{tab:pick-mouse-scene}
\end{table}

\begin{table}[ht]
  \centering
  \caption{Average success rates of SKIL and baselines on the \textit{Fold Towel} task under three types of environment changes. (\textit{Situation 0} means without any environmental changes.)}
  \begin{tabular}{@{}ccccc@{}}
    \toprule
    Method & \textit{Situation 0} & \textit{Situation 1} & \textit{Situation 2} & \textit{Situation 3} \\
    \midrule
    DP & 6/10 & 5/10 & 5/10 & 3/10 \\
    DP3 & 6/10 & 6/10 & 5/10 & 2/10 \\
    RISE & 5/10 & 4/10 & 2/10 & 4/10 \\
    GenDP-S & 6/10 & 5/10 & 4/10 & 4/10 \\
    SKIL (Ours) & \textbf{8/10} & \textbf{7/10} & \textbf{7/10} & \textbf{8/10} \\
    \bottomrule
  \end{tabular}
  \label{tab:fold-towel-scene}
\end{table}

\begin{table}[ht]
  \centering
  \caption{Average success rates of SKIL on the \textit{Hang Towel} task under three types of environment changes. (\textit{Situation 0} means without any environmental changes.)}
  \begin{tabular}{@{}ccccc@{}}
    \toprule
    Method & \textit{Situation 0} & \textit{Situation 1} & \textit{Situation 2} & \textit{Situation 3} \\
    \midrule
    Baselines & 0/10 & 0/10 & 0/10 & 0/10 \\
    SKIL (Ours) & \textbf{7/10} & \textbf{7/10} & \textbf{7/10} & \textbf{6/10} \\
    \bottomrule
  \end{tabular}
  \label{tab:hang-towel}
\end{table}

\subsection{Ablation Study}

\subsubsection{Ablation of Vision Foundation Models}
\label{app:vision-models}
We present the results of different foundation models in three tasks, \textit{Pick Mouse}, \textit{Fold Towel}, and \textit{Grasp Handle of Cup}. Table \ref{tab:dino-radio-dift-results} reports the success rate of the three tasks. The lowercase letters b, l, and g represent base, large, and giant, respectively, indicating different sizes of the ViT architecture. Similarly, the uppercase letters B, L, and H stand for Base, Large, and Huge, also denoting varying scales of the ViT architecture. Additionally, DiFT \cite{tang2023emergent} utilizes Stable Diffusion 2.1 as its vision foundation model.

\begin{figure*}[ht]
  \centering
  \includegraphics[width=\linewidth]{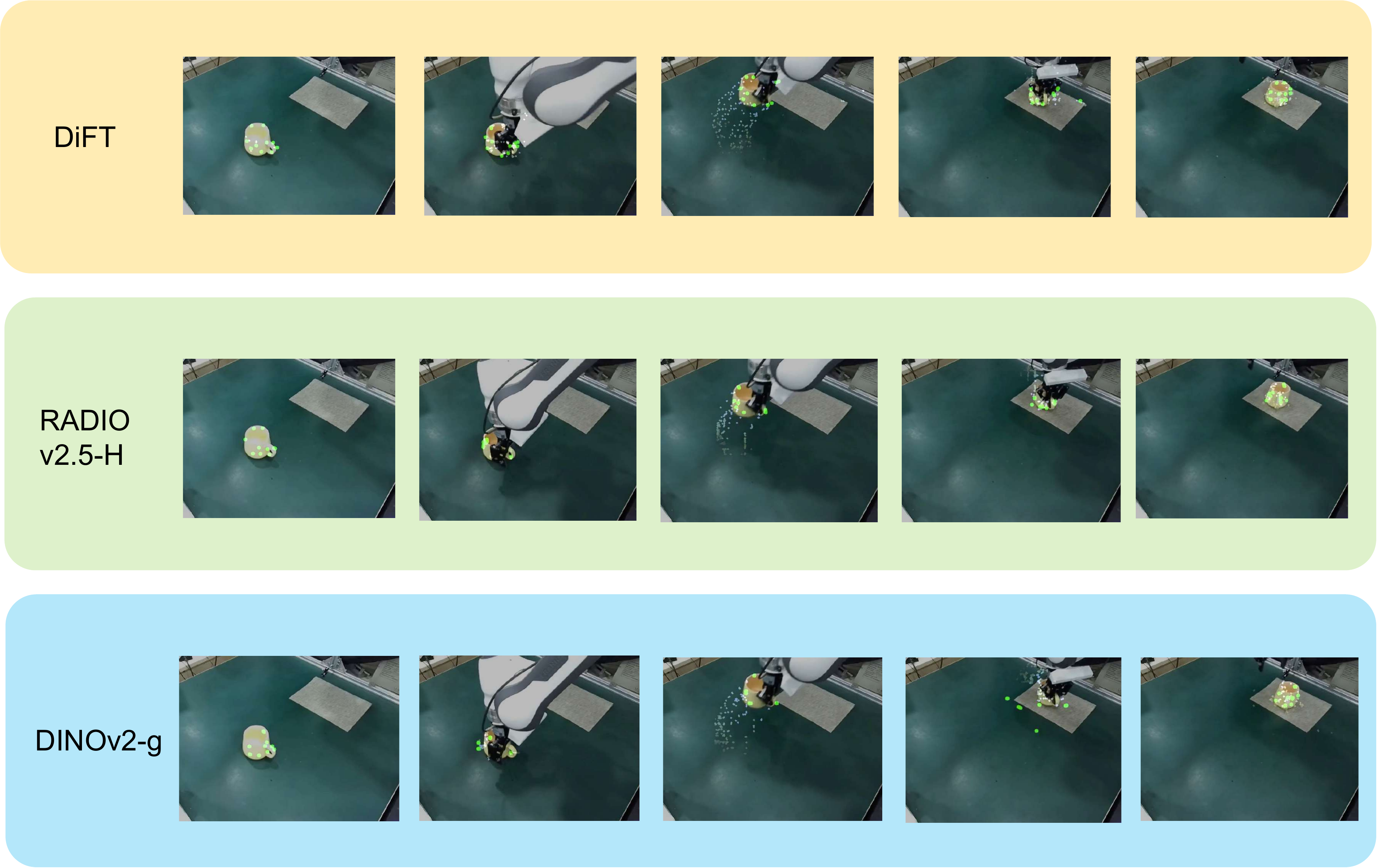}
  \caption{Visualization of keypoints' movement using three vision foundation models on \textit{Grasp Handle of Cup}.}
  \label{fig:kp-vlm}
\end{figure*}

Figure \ref{fig:kp-vlm} depicts the movement of semantic keypoints using different vision foundation models. We can see that the semantic keypoints obtained by DiFT \cite{tang2023emergent} and RADIO \cite{ranzinger2024radio} remain relatively accurate during the evaluation, while the keypoints from DINOv2 \cite{oquab2023dinov2} become detached from the relevant object's surface when meeting occlusions of keypoints.

\begin{table}[t]
  \centering
  \caption{Average success rates of various vision foundation models.}
  \begin{tabular}{@{}cccc@{}}
    \toprule
    Method & \textit{Pick Mouse} & \textit{Fold Towel} & \textit{Grasp Handle of Cup} \\
    \midrule
    DINOv2-b & 4/10 & 5/10 & 7/10 \\
    DINOv2-l & 4/10 & 6/10 & 6/10 \\
    DINOv2-g & 5/10 & 7/10 & 6/10 \\
    RADIOv2.5-B & 6/10 & 6/10 & 9/10 \\
    RADIOv2.5-L & 7/10 & \textbf{8/10} & 9/10 \\
    RADIOv2.5-H & 7/10 & 7/10 & 9/10 \\
    DiFT & \textbf{8/10} & 7/10 & \textbf{10/10} \\
    \bottomrule
  \end{tabular}
  \label{tab:dino-radio-dift-results}
\end{table}

\subsection{Details of Diffusion Action Head}
\label{app:diff_head}
We provide a detailed formulation of the diffusion model as follows. Note that we refer to some formulations in \cite{liu2024rdt}.

First, the denoising process is represented as:
$$
\mathbf{a}_t^{k-1} = \frac{\sqrt{\bar{\beta}^{k-1}} \gamma^k}{1 - \bar{\beta}^k} \mathbf{a}_t^0 + \frac{\sqrt{\beta^k}\left(1 - \bar{\beta}^{k-1}\right)}{1 - \bar{\beta}^k} \mathbf{a}_t^k + \tau^k \mathbf{v}
$$
Here, the parameters $\left\{ \beta^k \right\}_{k=1}^K$ and $\left\{ \tau^k \right\}_{k=1}^K$ are scalar coefficients from a predefined noise schedule. The terms are defined as $\gamma^k := 1 - \beta^k$ and $\bar{\beta}^{k-1} := \prod_{i=1}^{k-1} \beta^i$. Additionally, $\mathbf{v} \sim \mathcal{N}(\mathbf{0}, \mathbf{I})$ when $k > 1$; otherwise, $\bar{\beta}^{k-1} = 1$ and $\mathbf{v} = \mathbf{0}$.

The network is trained by minimizing the mean-squared error (MSE) between the predicted and true actions:
$$
\mathcal{L}(\boldsymbol{\phi}) := \operatorname{MSE}\left(\mathbf{a}_t, D_{\boldsymbol{\theta}} \left(\mathbf{o}_t, \sqrt{\bar{\beta}^k} \mathbf{a}_t + \sqrt{1 - \bar{\beta}^k} \mathbf{\epsilon}, k\right)\right)
$$
where $k \sim \operatorname{Uniform}(\{1, \dots, K\})$, $\mathbf{\epsilon} \sim \mathcal{N}(\mathbf{0}, \mathbf{I})$, and $(\mathbf{o}_t, \mathbf{a}_t)$ is sampled from the training dataset. For simplicity, noisy action inputs are denoted as $\tilde{\mathbf{a}}_t := \sqrt{\bar{\beta}^k} \mathbf{a}_t + \sqrt{1 - \bar{\beta}^k} \mathbf{\epsilon}$, where the index $k$ is omitted for clarity.




\subsection{Hyperparameters}~\label{app:train-detail}
We report the main hyperparameters for SKIL and SKIL-H in Table~\ref{tab:skil-hyperparameter} and Table~\ref{tab:skilh-hyperparameter}, respectively.
They share the same hyperparameters of the transformer encoders, as listed in Table \ref{tab:transformer-encoder-params}.
Following DP3 \cite{ze20243d}, we set the action prediction and execution horizon to be $H = 4$ and $N_{act} = 2$ in MetaWorld \cite{yu2020meta} and DexArt \cite{bao2023dexart}. 
We omit the diffusion model parameters in Table \ref{tab:skilh-hyperparameter} since the same parameters are used for both SKIL and SKIL-H. 
    
All models are trained on 2 NVIDIA 3090 GPUs, and the checkpoint with the lowest validation loss is saved as the final model for real-world performance evaluation. In simulation, we compute the average of the top three success rates every 100 training epochs.

\begin{table}[ht]
  \centering
  \caption{Main Parameters of the Transformer Encoder.}
  \begin{tabular}{@{}ll@{}}
    \toprule
    \textbf{Parameter}     & \textbf{Transformer Encoder}    \\ 
    \midrule
    Number of layers    & 1 \\
    Hidden size            &    128 \\
    Number of attention heads & 8  \\
    Feed-forward size       &  512         \\
    Dropout rate            &  0.1        \\
    Activation function   & ReLU      \\
    Attention type           & Self-Attention      \\
    Layer normalization     & Post-LN  \\
    Embedding size           & 128   \\
    Positional encoding     & Sinusoidal      \\
    \bottomrule
  \end{tabular}
  \label{tab:transformer-encoder-params}
\end{table}

\begin{table}[ht]
  \centering
  \caption{Hyperparameters of SKIL.}
  \begin{tabular}{@{}ll@{}}
    \toprule
    \textbf{Hyperparameters} & \textbf{SKIL} \\
    \midrule
    Epoch & 1000  \\
    Batch size &  256  \\
    Optimizer & AdamW  \\
    Learning rate & 1e-4  \\
    Weight decay & 1e-6 \\
    Lr scheduler & Cosine  \\
    \midrule
    \multicolumn{2}{c}{Diffusion Head in Policy Module} \\
    \midrule
    Noise scheduler & DDIM \\
    Denoising steps & 100(train); 10(test) \\
    Prediction horizon & 16 (4 in simulation) \\
    Observation horizon & 4 (2 in simulation)\\
    Action horizon & 8 (2 in simulation) \\
    \bottomrule
  \end{tabular}
  \label{tab:skil-hyperparameter}
\end{table}

\begin{table}[ht]
  \centering
  \caption{Hyperparameters of SKIL-H.}
  \begin{tabular}{@{}ll@{}}
    \toprule
    \textbf{Hyperparameters} & \textbf{SKIL-H}  \\
    \midrule
    Epoch & 1000  \\
    Batch size &  256  \\
    Optimizer & AdamW  \\
    Learning rate & 1e-4  \\
    Weight decay & 1e-6 \\
    Lr scheduler & Cosine  \\
    \midrule
    \multicolumn{2}{c}{Diffusion Head in Trajectory Prediction Module} \\
    \midrule
    Noise scheduler & DDIM \\
    Denoising steps & 100(train); 10(test) \\
    Prediction horizon  & 8  \\
    Observation horizon & 2 \\
    Action horizon & 4 \\
    \bottomrule
  \end{tabular}
  \label{tab:skilh-hyperparameter}
\end{table}


\end{document}